\pdfoutput=1
\documentclass[twoside]{article}
\usepackage[preprint]{aistats2027}
\usepackage[round]{natbib}

\usepackage[utf8]{inputenc}
\usepackage[T1]{fontenc}
\usepackage[hidelinks]{hyperref}
\usepackage{booktabs}
\usepackage{amsfonts}
\usepackage{amssymb}
\usepackage{amsmath}
\usepackage{nicefrac}
\usepackage{microtype}
\usepackage{graphicx}

\begin{document}

\runningtitle{What a Reporting Convention Hides}
\runningauthor{Lu Wei, Yufeng Wang, Haibin Ling}

\twocolumn[
\aistatstitle{What a Reporting Convention Hides: A Matched-Budget Audit of\\
Quantum Natural Gradient with an Exactly Computed Metric}
\aistatsauthor{Lu Wei$^{*}$ \And Yufeng Wang$^{*}$ \And Haibin Ling}
\aistatsaddress{Stony Brook University \And Stony Brook University \And Westlake University}
]
{\renewcommand{\thefootnote}{\fnsymbol{footnote}}\footnotetext[1]{Equal contribution.}}

\begin{abstract}
Several published comparisons of variational quantum optimizers time only runs that reach a target
loss, or read the verdict at a single target. Either convention could decide whether an optimizer's
costlier steps pay off. We measure how much each convention changes verdicts among Adam, simultaneous
perturbation stochastic approximation (SPSA) and quantum natural gradient (QNG), on initializations
held out from the selection of settings. We compute the exact metric that preconditions QNG, price every step
in circuit evaluations and give every method the same budget. In a median pooled over circuit widths,
cost families and a sweep of settings with common misses, SPSA needs more than twice Adam's
evaluations to reach a loose target. Dropping the censored runs that miss the target hides this gap.
On the global-cost family we hold fixed the settings selected for a strict target. QNG then usually
reaches the loose target after Adam but the strict target first. QNG's strict-target lead disappears
when the metric's simulator price, linear in the number of parameters, is replaced by an
assumed hardware count quadratic in that number. We recommend charging every miss the budget and
reporting verdicts across targets.
\end{abstract}

\section{Introduction}\label{sec:intro}

An optimizer that costs more per step is worth its price only if it reaches a useful loss with less
total computation. Benchmarks judge this by the time to target, the computation a run spends
before its loss first reaches a target level $\tau$. One such question is whether
the exactly computed preconditioner of quantum natural gradient (QNG) \citep{stokes2020}, the quantum
counterpart of the Fisher information, repays its price against Adam \citep{kingma2015} and against
simultaneous perturbation stochastic approximation (SPSA) \citep{spall1992}. SPSA's cheaper step
estimates the gradient from two loss values along one random direction.

\begin{figure}[t]
\centering
\includegraphics{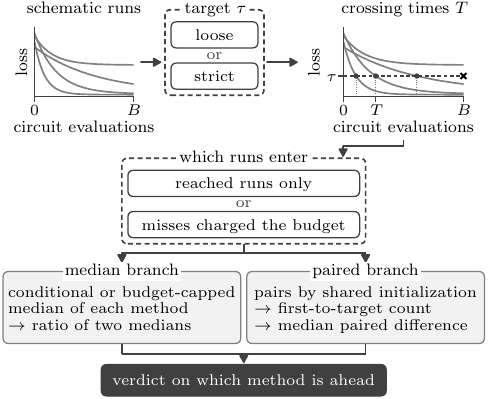}
\caption{The target $\tau$ defines the crossing times $T$, and $\times$ marks a run that does not reach $\tau$ within the budget $B$. The choice of which runs enter then acts on these times in both branches (Figure~\ref{fig:conventions}). Medians are conditional (reached runs only) or budget-capped (misses charged the budget). The paired branch counts which method needs fewer evaluations to the target, a miss counting as the budget, and takes the median of the paired differences.}
\label{fig:pipeline}
\end{figure}

Figure~\ref{fig:pipeline} traces two reporting choices that turn the runs of a time-to-target
comparison into a verdict. One is which runs enter the
statistic. A run that misses the target within its budget has a time right-censored at the budget, and
success-only timing drops such runs. This complete-case summary makes runtime estimates optimistic
\citetext{\citealp[Sec.~V-E]{mattos2021}; \citealp[Sec.~2.2]{tornede2020}}. A reading is a summary
computed under one option of a reporting choice. The two readings of which runs enter are
the conditional median over reached runs only and the budget-capped median, which charges every miss
the budget (Section~\ref{sec:setup}). The other choice is the target, which single-target
reporting fixes at one level. Authors of optimization benchmarks charge a miss instead of dropping it
and compare methods at several targets (Section~\ref{sec:related}). Several published comparisons of variational quantum
optimizers, however, time only the runs that reached a threshold, a target in our terms
\citep{wierichs2020, sung2020, jones2025}, or read their verdict at a single threshold \citep{fitzek2024}. To our knowledge, how much each choice changes such a verdict, with intervals and out of
sample, is open for variational quantum optimizers. If either choice can
change it, a claim that one optimizer is cheaper describes the optimizer and the reporting choice
together. We argue that, for optimizers whose steps differ
in price, the verdict depends on the target and on which runs enter the statistic, so one
number cannot report it. We measure this dependence on a testbed in which every step is charged an
explicit price in circuit evaluations and the preconditioner is computed exactly.

\begin{figure}[t]
\centering
\includegraphics{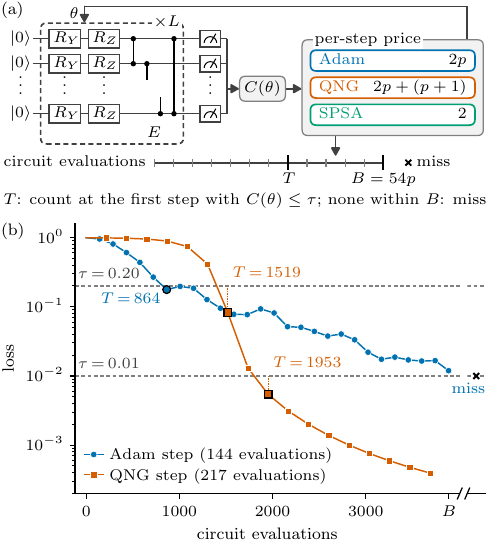}
\caption{(a)~The testbed. Each of the $L$ layers applies $R_Y$ and then $R_Z$ to every qubit and then the ring $E$ of controlled-$Z$ gates, and the loss $C(\theta)$ follows from the probabilities of measuring 0 (Section~\ref{sec:setup}). Each optimizer step adds its price in circuit evaluations to the run's count, $2p$ for Adam, $2p + (p + 1)$ for QNG and $2$ for SPSA, up to the budget $B = 54p$. (b)~Adam ($\eta = 0.3$) and QNG ($\eta = 1$, $\lambda = 10^{-1}$), with settings selected for $\tau = 0.01$, from one of the 20 new global-cost initializations at $n = 6$ ($B = 3888$) whose $\tau = 0.01$ paired difference is nearest the median (rule in Appendix~\ref{app:grid}). Adam is first at $\tau = 0.20$ and QNG at $\tau = 0.01$, which Adam misses within $B$. A black-edged marker is a run's first step at or below a target.}
\label{fig:pair}
\end{figure}

The models are variational
quantum circuits, and every cost is counted in circuit evaluations, one simulation of a circuit each.
Because the simulation is exact and noiseless, the only randomness is a run's starting point, or
initialization, and SPSA's random directions. Each loss, the global or the local cost of
\citet{cerezo2021}, lies between 0 and 1, so one target level serves both. A configuration is one cost family at one width of 3 to 6
qubits, which gives eight configurations. The global-cost family is prone to barren plateaus, in which the gradient
variance vanishes exponentially with the number of qubits \citep{mcclean2018, cerezo2021}.
Every step has an explicit integer price. A step costs $2p$ circuit evaluations for Adam with $p$ parameters, $2$ for SPSA and
$2p + (p + 1)$ for QNG, whose preconditioner is charged its simulator price $p + 1$. Every method receives the same budget of $B = 54p$
evaluations per run, and no wall-clock measurement enters the comparison (Section~\ref{sec:setup}; Figure~\ref{fig:pair}(a)).

The preconditioner is exact. In the classical deep-learning setting, curvature-based
optimizers are evaluated through an estimator, such as Kronecker factors, block-diagonal
approximations \citep{martens2015, gupta2018}, minibatch empirical Fisher, or staleness between refreshes. When such a
method underperforms, the loss cannot be attributed between the preconditioning idea and the
approximation of the curvature object. The quantum setting does not share that limitation, a fact that
prior work has exploited: \citet{wierichs2020} construct the Fubini--Study matrix exactly and
price it explicitly. QNG's preconditioner is the Fubini--Study metric $g$ of the state the circuit
prepares, computed in closed form. There is no
sampling, no factorization, and no damping schedule chosen to stabilize an estimate. QNG inverts
the damped metric $g + \lambda I$, with $I$ the identity and the damping $\lambda$ selected like the learning rate (Section~\ref{sec:setup}).

Which runs enter decides whether SPSA's gap to Adam at the loose target ($\tau = 0.20$) is visible in a median pooled over configurations. With every miss charged the budget, SPSA needs more than twice Adam's circuit evaluations in that median.
SPSA's gap to Adam replicates on 20 new initializations per configuration, held out from every selection made on the 10 original ones, and under SPSA's
per-initialization median of five sequences of random directions. Each replication has an adjusted bootstrap interval above~1
(Section~\ref{sec:audit}; row~1 of Table~\ref{tab:choices}; Figure~\ref{fig:conventions}(a)).

The target decides whether the exact metric repays its price. Every method starts from the same
initializations, so the runs of two methods form pairs, and the first of a pair is the method that
reaches the target with fewer circuit evaluations, a miss counting as the budget. With the settings
selected for the strict target ($\tau = 0.01$) held fixed on the global-cost family
(row~2 of Table~\ref{tab:choices}; Figure~\ref{fig:conventions}(b); Figure~\ref{fig:pair}(b) shows one
pair), a benchmark evaluated at the loose target concludes that the exact metric does not repay its
price. A benchmark evaluated at the strict target concludes that it does. Any two crossing loss curves
reverse their order at some pair of targets. On the global-cost family the change falls between
the loose and the strict target, which stand for the low- and high-accuracy levels that benchmarks
report side by side \citep{more2009, hansen2016}. Over six targets from 0.5 to 0.01, the estimated probability that QNG reaches the target first rises from about 0.04 to 0.96 (Section~\ref{sec:target}).
Table~\ref{tab:choices} also states when each verdict change is absent.

\begin{table*}[t]
\centering
\caption{Each row gives one comparison under two readings on its own fixed set of runs. Pooled medians and median paired differences (QNG minus Adam) mix budgets. Bootstrap intervals, with initializations resampled within configurations, are Bonferroni-adjusted 99.5\% (ten ratios per initialization set) in row~1 and 95\% in row~2. In row~2 first means fewer evaluations to $\tau$, a miss counting as the budget, and counts are QNG first--Adam first--tied. With each target's own settings, on different runs at the two targets, Adam is first from 66 of 80 at the loose and QNG from 77 at the strict target. Section~\ref{sec:setup} gives the definitions.}
\label{tab:choices}
\small\setlength{\tabcolsep}{4pt}
\begin{tabular}{@{}p{0.43in}p{1.60in}p{0.78in}p{0.78in}p{2.717in}@{}}
\toprule
\raggedright Choice &
\raggedright Comparison; held fixed &
\raggedright Reading 1 &
\raggedright Reading 2 &
\raggedright Verdict change absent when \tabularnewline
\midrule
\raggedright Which runs enter (Sec.~\ref{sec:audit}) &
\raggedright SPSA vs.\ Adam, ratio of median times pooled over eight configurations; original grid and initializations, loose target, SPSA's shared perturbation sequence &
\raggedright Reached runs only\newline 1.11 (0.84--1.40) &
\raggedright Misses charged the budget\newline 2.24 (1.60--2.69) &
\raggedright Settings selected for the loose target (every Adam and SPSA run reaches it on both initialization sets, so the readings coincide) or one configuration taken alone (Reading~1 still shrinks the ratio but leaves SPSA slower in 7 of 8) \tabularnewline
\midrule
\raggedright At which target (Sec.~\ref{sec:target}) &
\raggedright QNG vs.\ Adam, number of initializations with QNG first and median paired difference; global-cost family, 80 new initializations, strict-target settings, simulator price &
\raggedright Loose target, $\tau = 0.20$\newline 12 of 80\newline $+172$ (121.5--255) &
\raggedright Strict target, $\tau = 0.01$\newline 77 of 80\newline $-958.5$ ($-1099.5$ to $-842.5$) &
\raggedright Settings selected for the loose target, evaluated at the strict target (40--29--11, sign-test p-value 0.23; Adam first at 5 and 6 qubits, 2--14--4 and 1--12--7), or the metric charged $p(p+1)/2$ evaluations per step (QNG first from 0 of 80 at the loose and 3~of 59 non-tied at the strict target) \tabularnewline
\bottomrule
\end{tabular}
\end{table*}%

Addressing the above issues, in this paper we make the following main contributions:
\begin{itemize}
\item We show, with intervals on two sets of initializations, that success-only timing hides a more
than twofold gap between SPSA and Adam in a median pooled over a sweep with common misses, and shrinks
QNG's gap to Adam without hiding it (Section~\ref{sec:audit}).
\item We show that the verdict between exact-metric QNG and Adam on the global-cost family reverses
between the loose and the strict target, with settings fixed and tested out of sample, and we measure
how far it depends on the selection target and on the metric's price (Section~\ref{sec:target}).
\item We construct a paired testbed that charges every step an explicit price in circuit evaluations
and in which a longer budget cannot change any QNG--Adam first-to-target count of the main
comparison (Section~\ref{sec:setup}), and we describe the exact metric's spectrum as depth grows
(Section~\ref{sec:ladder}).
\end{itemize}
We recommend that a time-to-target comparison of optimizers whose steps differ in price report its
verdict as a curve over targets and charge every miss the budget (Section~\ref{sec:report}).

\section{Related work}\label{sec:related}

Rankings of deep-learning optimizers change with the search space \citep[Abstract]{choi2019} and the tuning budget \citep[Fig.~1]{sivaprasad2020}. A non-diagonal preconditioner won the AlgoPerf external-tuning track \citep{kasimbeg2025}, so whether a costly curvature step pays is a live question. Fair comparisons of optimizers require equal tuning effort \citetext{\citealp[Sec.~4.4]{beiranvand2017}; \citealp[p.~359]{birattari2006}}, settings selected on instances other than those that score them \citep[Sec.~2]{birattari2006} and a price for every step \citep[Sec.~2.3]{schneider2019}, such as the per-step circuit evaluations of \citet[Table~1]{kverne2026}. We use all three as controls in Sections~\ref{sec:setup} and~\ref{sec:target}.
Optimization benchmarks also charge a run that misses its target instead of dropping it \citep{dolan2002, more2009, hoos1998, hansen2016, dahl2023} and compare methods at several targets \citep{more2009, hansen2016, hoos2008}, at which the better algorithm can differ \citep[Sec.~5]{bartzbeielstein2020}. We measure how much each of these choices changes a verdict.

Natural-gradient descent preconditions the gradient with the Fisher information \citep[Sec.~3.1]{amari1998}, which deep-learning optimizers estimate (Section~\ref{sec:intro}), for instance by the empirical Fisher \citep{kunstner2019}. QNG uses the Fubini--Study metric \citep{stokes2020}, whose full form needs $O(p^2)$ circuit evaluations on hardware \citep[Sec.~3.1]{gacon2021}. \citet{wierichs2020} build and price the full metric and weigh the reliability of natural gradient against its cost (Appendix~\ref{app:related}). Relative to that study, we match the evaluation budget, select settings on
separate initializations, include SPSA and report intervals.

Some benchmarks of variational quantum optimizers compare final losses at a fixed budget \citetext{\citealp[Secs.~III and VI]{bonetmonroig2023}; \citealp[Sec.~6]{lavrijsen2020}}. Others time only the runs that reached the target, beside success rates or counts that show reliability but not cost \citep{wierichs2020, sung2020, jones2025}. Comparisons of QNG with Adam use approximate metrics or untuned step sizes or count iterations \citetext{\citealp[Sec.~3]{stokes2020}; \citealp[Sec.~3]{fitzek2024}; \citealp[Fig.~5(a)]{haug2021b}}. Two of them find the order of QNG and Adam changing during the runs, in opposite directions \citetext{\citealp[Sec.~3 and Table~1]{fitzek2024}; \citealp[Fig.~5(a)]{haug2021b}}. To our knowledge, no variational-algorithm benchmark computes one verdict under both readings of which runs enter, and none, including the multi-cutoff study of \citet{sung2020}, tests a change of order between two targets on paired runs, with settings selected on separate initializations and intervals at both targets (Appendix~\ref{app:related}).
Barren plateaus, to which the global-cost family is prone, also suppress Hessian entries \citep{cerezo2021b} and the loss differences of gradient-free optimizers \citep{arrasmith2021}. In quantum machine learning, \citet{bowles2024} argue that experimental-design choices strongly shape simulated benchmarks of quantum models. We apply the same scrutiny to benchmarks of optimizers.

\section{Setup and statistics}
\label{sec:setup}

Two cost families on one hardware-efficient circuit \citep{kandala2017} give landscapes whose gradient variance falls with width at different rates.
The state of $n$ qubits is a unit vector $|\psi\rangle \in \mathbb{C}^{2^n}$, and $\langle\psi|$ is its conjugate transpose, so $\langle\phi|\psi\rangle$ is the inner product of two states $|\phi\rangle$ and $|\psi\rangle$.
With $|0\rangle = (1, 0)^{\top}$ and $|1\rangle = (0, 1)^{\top}$, the basis vectors are the Kronecker products $|x\rangle = |x_1\rangle \otimes \cdots \otimes |x_n\rangle$ for bit strings $x \in \{0, 1\}^n$, and $|0\cdots0\rangle = |0\rangle^{\otimes n}$.
With $\mathrm{i}^2 = -1$ and the Pauli matrices $Y = \bigl(\begin{smallmatrix}0 & -\mathrm{i}\\ \mathrm{i} & 0\end{smallmatrix}\bigr)$ and $Z = \bigl(\begin{smallmatrix}1 & 0\\ 0 & -1\end{smallmatrix}\bigr)$, the rotations by an angle $\alpha$ are $R_Y(\alpha) = e^{-\mathrm{i}\alpha Y/2}$ and $R_Z(\alpha) = e^{-\mathrm{i}\alpha Z/2}$. The operator $R^{(q)}$ applies a rotation $R$ to qubit $q$ only.
The controlled-$Z$ gate on qubits $q$ and $q'$ multiplies by $-1$ every $|x\rangle$ with $x_q = x_{q'} = 1$. The ring $E$ applies it to the pairs $(1, 2), \dots, (n - 1, n)$ and $(n, 1)$.
Layer $\ell$ applies $R_Y(\theta_{\ell q 1})$ and then $R_Z(\theta_{\ell q 2})$ to every qubit $q$ and then $E$, and the circuit has $L$ layers (Figure~\ref{fig:pair}(a)):
\begin{align}
U_\ell &= E\, \textstyle\prod_{q=1}^{n} R_Z^{(q)}(\theta_{\ell q 2})\, R_Y^{(q)}(\theta_{\ell q 1}), \label{eq:circuit}\\
|\psi(\theta)\rangle &= U_L \cdots U_1\, |0\cdots0\rangle, \nonumber
\end{align}
so the circuit has $p = 2nL$ parameters, the angles collected in $\theta$ and indexed by $i$.
Measuring every qubit returns $x$ with probability $\lvert\langle x|\psi(\theta)\rangle\rvert^{2}$.
The global-cost family has depth $L = n$ and the loss $C(\theta) = 1 - \lvert\langle 0 \cdots 0|\psi(\theta)\rangle\rvert^{2}$, one minus the probability that all $n$ qubits are measured as $0$. The local-cost family has depth $L = 2$ and the loss $C(\theta) = 1 - \frac{1}{n}\sum_{j=1}^{n} \pi_j(\theta)$, where $\pi_j(\theta)$ is the probability that qubit $j$ is measured as $0$. The two losses are those of \citet{cerezo2021}, and the circuit and depths are our choice.
The widths $n \in \{3, 4, 5, 6\}$ give $p$ from 12 to 72.
The log variance of one partial derivative has fitted slopes in $n$ of $-1.41$ to $-1.35$ for the global-cost and $-0.54$ to $-0.51$ for the local-cost family, all with 95\% bootstrap intervals excluding zero (Appendix~\ref{app:bp}).

Each optimizer step is charged an explicit price in circuit evaluations, and the per-run budget is fixed for each configuration before any run.
Because every angle enters through one rotation, the parameter-shift rule gives $\partial C/\partial\theta_i = \tfrac{1}{2}[C(\theta + \tfrac{\pi}{2}e_i) - C(\theta - \tfrac{\pi}{2}e_i)]$, with $e_i$ the $i$th unit vector \citep[eq.~(3)]{mitarai2018}.
Here the Fisher-type preconditioner is the Fubini--Study metric
$g_{ij} = \mathrm{Re}\big[\langle\partial_i\psi|\partial_j\psi\rangle -
\langle\partial_i\psi|\psi\rangle\langle\psi|\partial_j\psi\rangle\big]$, where
$|\psi\rangle$ is the state the circuit prepares at parameters $\theta$ and
$|\partial_i\psi\rangle$ its derivative with respect to $\theta_i$.
It is computed in closed form by analytic generator insertion.
The state $|\partial_i\psi\rangle$ is computed as the output of the circuit with $-\tfrac{\mathrm{i}}{2}Y$ or $-\tfrac{\mathrm{i}}{2}Z$ inserted after rotation $i$, so $g$ needs $p + 1$ states.
QNG takes the step $\delta\theta = -\eta\,(g + \lambda I)^{-1}\nabla C$, the natural-gradient update of \citet[eqs.~12--13]{stokes2020} with the Tikhonov damping of \citet[Sec.~II.B.3]{wierichs2020}. Here $\lambda$ is the damping, $I$ the $p \times p$ identity and $\nabla C$ the gradient of the loss, and $\eta$ is the learning rate, which Adam \citep{kingma2015} and SPSA \citep{spall1992} also have.
With $m$ the price of one metric, the prices per step are
\begin{equation}
\text{Adam: } 2p, \qquad \text{QNG: } 2p + m, \qquad \text{SPSA: } 2,
\label{eq:prices}
\end{equation}
since a partial derivative costs two evaluations and SPSA uses two loss values per step.
Hardware pays this gradient price, although a simulator could obtain the gradient from the $p + 1$ states of the metric.
The metric is charged its simulator price $m = p + 1$, one evaluation per state and the lowest price considered; on hardware it costs more \citep{gacon2021}.
No method is charged for the loss recorded after each step, and charging it changes neither finding of Sections~\ref{sec:audit} and~\ref{sec:target} (Appendix~\ref{app:budget}).
The budget $B = 54p$, from 648 to 3888 evaluations, allows 27 Adam steps, 17 QNG steps and $27p$ SPSA steps.
QNG's 17 steps spend 94.9--97.1\% of $B$.
A higher price $m$ never raises the number of initializations from which QNG reaches a target before Adam (Appendix~\ref{app:price}).

Optimizer comparisons are evaluated at the loose target $\tau = 0.20$ or the strict target $\tau = 0.01$.
A run is one method at one setting, $\eta$ and for QNG also $\lambda$, started from one initialization and stopped when its budget is spent. Its final loss is the loss after its last step.
Its time to target $T$, or crossing time, is the cumulative number of circuit evaluations at the end of the first step with loss at or below $\tau$. A run whose initial loss is already at or below $\tau$ has $T = 0$ (Appendix~\ref{app:zero}).
A run that does not reach $\tau$ within $B$ is a miss, whose time is known only to exceed $B$ and is censored there.
The $p$ angles of an initialization are drawn independently and uniformly from $[0, 2\pi)$.
The \emph{original grid}, a sweep of settings not selected for any target, is a two-point learning-rate grid with two settings each for Adam and SPSA, $\eta \in \{0.1, 0.3\}$. For QNG it adds $\lambda \in \{10^{-3}, 10^{-1}\}$ and has four settings.
On the 10 \emph{original initializations} of each configuration, which serve for selection, it gives 640 runs, 160 for Adam, 320 for QNG and 160 for SPSA.
Each method's \emph{best original setting} is $\eta = 0.3$, with $\lambda = 10^{-3}$ for QNG. That setting has the lowest median of $\min(T, B)$ on the original grid in every configuration at both targets, with ties resolved as in Appendix~\ref{app:grid}.
The \emph{extended grid}, the search space for selection on the same initializations, has six learning rates from 0.03 to 10 for Adam and SPSA and 27 settings for QNG, with $\lambda$ from $10^{-5}$ to $1$. On it we select, for each method, configuration and target, the setting with the lowest median of $\min(T, B)$ at that target, with ties broken as in Appendix~\ref{app:grid}.
Every original and every selected setting is then run from 20 \emph{new initializations} per configuration, held out from every selection.
The strict target was fixed after we examined six targets from 0.5 to 0.01 on the 640 original-grid runs and before any run on the new initializations.
The \emph{main comparison} of Section~\ref{sec:target} uses the strict-target selections on the new global-cost initializations, held fixed, with every miss charged the budget and the metric charged its simulator price.
Runs of the strict-target settings at $2B$ and $4B$ extend those at $B$, and no QNG--Adam pair at these settings has both runs missing or is tied at $B$. The first method therefore cannot change with the budget, while final loss can (Appendix~\ref{app:budget}).

Extending the grids where a selection fell at an edge moves 14 of the 24 local-cost selections from $\eta = 1$ to $\eta = 3$ and no global-cost selection (Appendix~\ref{app:grid}).
In every SPSA run described so far, all initializations of a configuration and learning rate share one perturbation sequence of random directions, SPSA's \emph{shared sequence} (Appendix~\ref{app:spsa}).
Every SPSA setting is also run under five independent sequences per initialization, and its per-initialization median of five sequences is the third smallest of the five crossing times, a miss being infinite.
Table~\ref{tab:app-protocol} lists every set of runs described above.

For $N$ runs with times $T_i$ and budgets $B_i$, the reach rate $r$, the conditional median $M_{\mathrm{cond}}$, the budget-capped median $M_{\mathrm{cap}}$ and the restricted mean time to target (RMST) are
\begin{align}
r &= \tfrac{1}{N}\textstyle\sum_{i} \mathbf{1}\{T_i \le B_i\}, \label{eq:reach}\\
M_{\mathrm{cond}} &= \operatorname{median}\{T_i : T_i \le B_i\}, \label{eq:cond}\\
M_{\mathrm{cap}} &= \operatorname{median}\{\min(T_i, B_i)\}, \label{eq:cap}\\
\mathrm{RMST} &= \tfrac{1}{N}\textstyle\sum_{i} \min(T_i, B_i)/B_i, \label{eq:rmst}
\end{align}
where $\mathbf{1}\{\cdot\}$ is 1 when its condition holds and 0 otherwise. The median of an even number of values is the midpoint of the two central ones.
Within a configuration every run has the same budget $B$, at which every miss is censored.
There $B \cdot \mathrm{RMST}$ is the restricted mean survival time up to $B$ \citep{royston2013}. It equals the area under the Kaplan--Meier curve \citep{kaplan1958} up to $B$ and the PAR-1 score of algorithm benchmarking \citep[Sec.~2.2, eq.~(1)]{wang2022}.
For the same reason $M_{\mathrm{cap}}$ equals the Kaplan--Meier median under the midpoint convention of Appendix~\ref{app:conv} when at least $N/2 + 1$ of the $N$ runs reach the target. When at most half do, it reflects the budget rather than the method.
A median pooled over configurations mixes budgets and need not equal a Kaplan--Meier median (Appendix~\ref{app:pool}).

\begin{figure*}[t]
\centering
\includegraphics{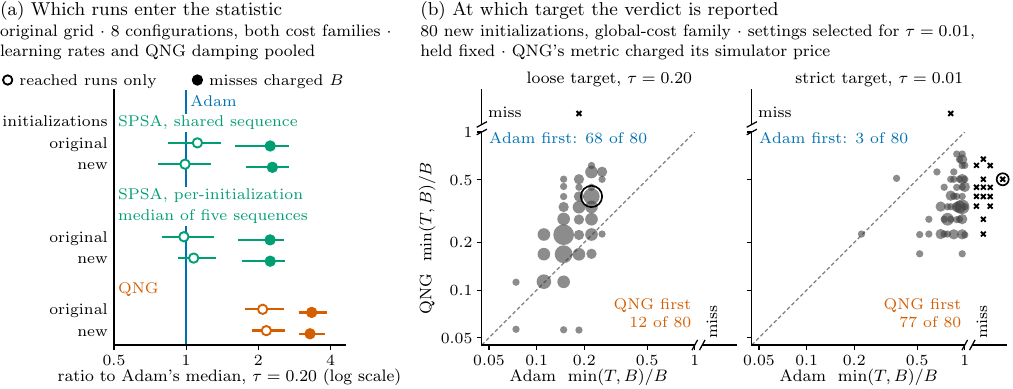}
\caption{Two reporting choices, each applied to its own fixed set of runs. Only Adam's $\eta = 0.3$
runs on the new global-cost initializations enter both panels (Appendix~\ref{app:grid}).
(a)~Original grid at $\tau = 0.20$, original and new initializations in separate rows, all settings
and the eight configurations of both cost families pooled, so medians mix budgets. Open markers use
reached runs only, and filled ones charge every miss $B$. The line at 1 is Adam, whose runs all
reach the target. SPSA's shared sequence is defined in Section~\ref{sec:setup}. Bars are
Bonferroni-adjusted 99.5\% bootstrap intervals, ten ratios per initialization set. (b)~Global-cost
family, 80 new initializations, settings selected for $\tau = 0.01$, held fixed. A grey marker
merges initializations with the same Adam and the same QNG step count, each displaced by at most
1\%, area proportional to their number. At Adam's value 1, Adam reached the target at its last step.
A cross in a strip is a miss, placed at the other run's value, and position across strips carries no
value. \emph{First} means fewer evaluations to the target, a miss counting as $B$, and counts
include the strips. The metric is charged its simulator price. Settings selected for $\tau = 0.20$
give other counts (Section~\ref{sec:target}). The circled marker is the pair of
Figure~\ref{fig:pair}(b).}
\label{fig:conventions}
\end{figure*}%

Because the methods share initializations, two methods are also compared pair by pair.
From one initialization, the first method is the one that reaches the target with fewer circuit evaluations, a miss counting as the budget, and equal counts are a tie.
For methods $a$ and $b$ we report the number of initializations from which each is first and the median of the paired differences $D_i = \min(T^{a}_i, B) - \min(T^{b}_i, B)$, positive when $a$ is slower.
The tests are the exact two-sided sign test on the non-tied pairs and the Wilcoxon signed-rank test, and a p-value is reported only when each method runs at one setting per configuration.
The two sign tests of Section~\ref{sec:target} at the fixed strict-target selection carry the main claim, and every other p-value is per-comparison and unadjusted.
Intervals are 95\% bootstrap intervals over initializations resampled within configurations, Bonferroni-adjusted for the ratio families of Appendix~\ref{app:conv}.

The parameter-shift derivatives and the full metric agree with central finite differences, and the gradient-variance estimate and the one-rotation metric agree with closed forms (Appendix~\ref{app:bp}).
Adam reaches the loose target in every original-grid run, so that target is attainable within $B$ from every initialization. The loose-target misses of QNG and SPSA therefore describe those methods at these settings and this budget.
Adam reaches the strict target in only 70.6\% of these runs.

\section{Success-only timing}
\label{sec:audit}

In a median pooled over configurations at the loose target, success-only timing hides SPSA's gap to Adam and shrinks QNG's gap without hiding it.
Over reached runs only, SPSA's median is 1.11 times Adam's, and the adjusted interval of this ratio covers~1.
With every miss charged the budget SPSA's median is 2.24 times Adam's (adjusted intervals in row~1 of Table~\ref{tab:choices}).
Both readings use the original-grid runs on the original initializations, with SPSA under its shared sequence, and differ only in which runs enter (Figure~\ref{fig:pipeline}; Table~\ref{tab:main} in Appendix~\ref{app:cells}).
Every Adam run reaches the loose target, so Adam's two medians coincide, and a ratio to Adam changes between the readings only through the other method's misses.

The pooled conditional median hides SPSA's gap because SPSA's misses concentrate in the two global-cost configurations with the largest budgets and the longest times.
SPSA's median over reached runs therefore draws on configurations with shorter times than Adam's median does (Appendix~\ref{app:cells}).
Within each configuration, success-only timing still shrinks SPSA's ratio to Adam but leaves SPSA slower than Adam in 7 of the 8 configurations on the original initializations and in 6 on the new ones (Table~\ref{tab:app-gap} in Appendix~\ref{app:cells}).
On the original initializations the exception is the local-cost configuration at $n = 3$. There the two readings order Adam and SPSA oppositely only when both learning rates are pooled and, under SPSA's shared sequence, only for the midpoint median (Appendix~\ref{app:cells}).
Dividing each time by its configuration's budget before pooling removes the mixing of budgets, and success-only timing still about halves SPSA's pooled ratio (Appendix~\ref{app:cells}).

SPSA's hidden gap replicates on the new initializations and under the per-initialization median of five sequences, which adds SPSA's own randomness.
In each case the conditional interval covers~1 and the budget-capped interval lies above it (Figure~\ref{fig:conventions}(a)).
Appendix~\ref{app:cells} gives the ratios on the new initializations and the factors by which success-only timing shrinks them, and Table~\ref{tab:spsa} in Appendix~\ref{app:spsa} gives SPSA's ratios under each perturbation sequence.

For QNG, success-only timing shortens the reported time without hiding the gap to Adam.
On the original grid and initializations, dropping QNG's misses lowers its pooled median by a factor of 1.60, and QNG's adjusted intervals lie above~1 under both readings on both initialization sets (Figure~\ref{fig:conventions}(a); Appendix~\ref{app:cells}).
Even at its best original setting, QNG's pooled budget-capped median at the loose target on the original initializations is 3.53 times Adam's (Appendix~\ref{app:cells}).

A failure rate also depends on the settings it pools over.
On the original grid and initializations QNG misses the loose target in 41.3\% of its runs, more often than SPSA under its shared sequence.
This rate pools four QNG settings against two for each other method. At its smaller damping, QNG misses less often than SPSA (Appendix~\ref{app:cells}).
At its best original setting QNG misses in 7.50\% of its runs, while every Adam run reaches the target (Appendix~\ref{app:cells}).

At the loose target the difference between the two readings shrinks as selection makes misses rare.
For QNG on the original initializations, the ratio of the budget-capped to the conditional median is smaller at its best original setting than over the original grid.
Under the loose-target selection every QNG run on these initializations reaches the target, and the two medians coincide (Appendix~\ref{app:cells}).
Which runs enter therefore matters in a median pooled over a sweep with common misses, as on the original grid. It does not matter once the settings are selected for the loose target, where both readings put SPSA ahead of Adam (Appendix~\ref{app:spsa}).

\section{Single-target reporting}
\label{sec:target}

With the settings selected for the strict target held fixed, the verdict between exact-metric QNG and Adam on the global-cost family reverses between the two targets. Adam reaches the loose target first from 68 of the 80 new initializations, and QNG reaches the strict target first from 77 (Figure~\ref{fig:conventions}(b)). Both exact sign tests give $p < 10^{-9}$ (Table~\ref{tab:app-select} in Appendix~\ref{app:select}). Where the first method changes between the targets, it changes from Adam to QNG, never the other way (Appendix~\ref{app:select}). Both counts come from the main comparison (Section~\ref{sec:setup}) and differ only in the target at which it is evaluated (Figure~\ref{fig:pipeline}). On the original grid and initializations the loose target supports the verdict that the exact metric does not repay its price. Adam reaches the loose target first in every non-tied pair in which the two methods share a learning rate, at each of QNG's four original settings, with both cost families pooled (Appendix~\ref{app:cells}).

Counted pair by pair, these first-to-target fractions estimate the probability, over initializations in the four global-cost configurations, that QNG reaches the target first. \citet[Sec.~4.1]{bouthillier2021} recommend this probability, rather than a comparison of single results, as the criterion for deciding between two methods whose results vary with random choices. Over the six examined targets at the same settings, the majority switches from Adam to QNG once on each family, between $\tau = 0.1$ and $0.05$ on the global-cost family and one target looser on the local-cost family. QNG's count rises at every stricter target in both families, and Table~\ref{tab:app-targets} in Appendix~\ref{app:select} gives each count with its sign test. The direction is the same among the global-cost pairs in which both runs reach the target (Appendix~\ref{app:select}).

The change also holds with each target's own selection and when QNG searches as many settings as Adam, and selection adds little optimism (Appendix~\ref{app:grid}). With each method at the setting selected for the target at which it is evaluated, QNG's RMST on the new global-cost initializations exceeds Adam's at the loose target and falls below it at the strict target, with both 95\% intervals excluding zero (Appendix~\ref{app:attain}). These comparisons use different runs at the two targets. When QNG's search over 27 settings for the strict target is restricted to Adam's six learning rates at the damping $\lambda = 10^{-3}$, it selects the best original setting in every global-cost configuration. QNG is then first at the strict target from 73 of the 80 new initializations and Adam at the loose target from 79 (Table~\ref{tab:app-select}). The original initializations, on which that setting was selected, show the same reversal (Appendix~\ref{app:cells}).

The target that selected the settings and the price charged for the metric bound the reversal. The settings selected for the loose target instead lower only QNG's damping at global-cost $n = 5$ and $6$ (Appendix~\ref{app:grid}). With them QNG's lead at the strict target over the 80 new global-cost initializations is no longer significant. Adam is first in those two configurations, and QNG stays first in the other two, where both targets select the same settings (row~2 of Table~\ref{tab:choices}; Appendix~\ref{app:select}).

The price bounds the reversal through the break-even price $m^\ast$, the largest per-step metric charge at which QNG's budget-capped median at the strict target stays below Adam's. With the strict-target settings on the new global-cost initializations, $m^\ast$ lies between $3.8p$ and $8.9p$ evaluations. An assumed hardware count of $p(p+1)/2$ charges one circuit evaluation per independent entry of the symmetric metric. It exceeds every bootstrap replicate of $m^\ast$ except at $n = 3$, where it lies inside the 95\% interval of $m^\ast$ (Appendix~\ref{app:price}). Under that count Adam is first at the strict target from almost every non-tied initialization (row~2 of Table~\ref{tab:choices}).

On the local-cost family the paired direction also reverses with each target's own selection, on different runs at the two targets, but its loose-target side rests on the sign test alone (Appendix~\ref{app:select}). A second protocol with one fixed setting per method repeats the strict-target reversal only on this paper's global-cost family, not on its local-cost family or on a parameter-tied circuit (Appendix~\ref{app:prot2}). Against SPSA, QNG's strict-target verdict depends on the test (Appendix~\ref{app:spsa}).

\section{Spectrum of the metric}
\label{sec:ladder}

\begin{table}[t]
\centering
\caption{The exact metric $g$ for the global cost at $n = 6$ is measured apart from the optimizer runs, not
per run, at 20 random initial points per depth; entries are means, without an uncertainty estimate.
The true rank counts the eigenvalues ${\lambda_{\max} \ge \dots \ge \lambda_{\min}}$ of $g$ above
$10^{-10}\lambda_{\max}$ and equals $p$, so every point is full rank and ${\kappa(g) =
\lambda_{\max}/\lambda_{\min}}$. Effective rank is $\operatorname{tr}(g)/\lambda_{\max}$; alignment
is ${\cos(\nabla C, (g + \lambda I)^{-1}\nabla C)}$ at damping ${\lambda = 10^{-3}}$.}
\label{tab:spectrum}
\small\setlength{\tabcolsep}{4pt}%
\begin{tabular}{@{}lrrrrrr@{}}
\toprule
Depth $L$ & 1 & 2 & 3 & 4 & 5 & 6 \\
\midrule
$p = {}$true rank & 12 & 24 & 36 & 48 & 60 & 72 \\
$\log_{10}\kappa(g)$ & 1.87 & 3.64 & 4.05 & 4.89 & 5.44 & 5.22 \\
Effective rank & 8.9 & 8.0 & 9.2 & 9.5 & 9.6 & 10.5 \\
Alignment & 1.000 & 0.434 & 0.436 & 0.330 & 0.301 & 0.294 \\
\bottomrule
\end{tabular}
\end{table}%

Along a depth ladder of the global-cost circuit at $n = 6$ with $L = 1, \dots, 6$ layers and $p = 12$ to $72$, the condition number of the metric $g$, whose damped form QNG inverts, rises by more than three decades while its effective rank stays between $8.0$ and $10.5$ (Table~\ref{tab:spectrum}; Appendix~\ref{app:ladder}). The condition number is $\kappa(g) = \lambda_{\max}/\lambda_{\min}$, where $\lambda_{\max}$ and $\lambda_{\min}$ are the largest and smallest eigenvalues of $g$, distinct from the damping $\lambda$. The effective rank is $\operatorname{tr}(g)/\lambda_{\max}$. The true rank is the number of eigenvalues above $10^{-10}\lambda_{\max}$. Averaged over 20 random initial points, $\log_{10}\kappa(g)$ rises from $1.87$ at $L = 1$ to $5.44$ at $L = 5$ and $5.22$ at $L = 6$, and the true rank equals $p$, below the ceiling $2(2^n - 1)$ at which \citet[Sec.~V]{haug2021} find this rank saturating as layers are added. No measurement here tests a link between this spectrum and either verdict.

\section{Discussion}
\label{sec:report}

A time-to-target comparison of optimizers whose steps differ in price should report its verdict as a curve over targets, with paired tests or intervals and the target at which the verdict reverses. It should give the reach rate beside every time and charge every miss the budget. It should also name the target that selected the settings, the price charged for each step, and the hyperparameter sweep and the configurations that each rate and median pools over. A Kaplan--Meier median or a restricted mean summarizes the censored times without dropping them, and a paired count estimates the probability that one method reaches the target first (Sections~\ref{sec:setup} and~\ref{sec:target}).

All results come from exact-gradient, noiseless simulation at $n \le 6$ and $p \le 72$, on one circuit family with two standard cost functions \citep{cerezo2021}. SPSA is designed for the shot-noise regime that this study excludes, and its standing may differ there. Block-diagonal, lazily refreshed and finite-shot approximations of the metric lie outside the reported evidence, so the findings concern this QNG algorithm at these prices. They do not show that the exactness of the metric causes either verdict. The new initializations test the selected settings out of sample within the eight configurations studied, not on other circuit families or at larger widths. Changing the optimizer does not avoid barren plateaus asymptotically, because a loss-minimizing direction still requires exponentially many measurements \citep{larocca2025}, and no result here implies such a mitigation. Shot noise and a measured hardware cost of the metric are the next tests of these verdicts.

\section{Conclusion}
\label{sec:conclusion}

We find that, on these runs, each reporting choice can reverse a verdict under the conditions in Table~\ref{tab:choices}. Which runs enter a pooled median decides whether SPSA's gap to Adam at the loose target is visible, unless the settings are selected for that target (Section~\ref{sec:audit}). On the global-cost family the target decides whether the exact metric repays its price, within the bounds set by the selection target and the metric price (Section~\ref{sec:target}).

\bibliographystyle{plainnat}
\bibliography{references}

\clearpage
\appendix
\onecolumn
\makeatletter
\newcommand{\fb@pending}{\begingroup\let\@elt\relax
  \xdef\fb@list{\@botlist\@deferlist\@dbldeferlist}\endgroup}
\newcommand{\FloatBarrier}{\par\fb@pending
  \ifx\fb@list\@empty\else\newpage\fb@pending\ifx\fb@list\@empty\else\clearpage\fi\fi}
\let\fb@xfloat\@xfloat\def\@xfloat#1[#2]{\fb@xfloat{#1}[htb]}
\makeatother
\let\appsection\section
\renewcommand{\section}[1]{\FloatBarrier\appsection{#1}\suppressfloats[t]}
\renewcommand{\topfraction}{0.95}\renewcommand{\bottomfraction}{0.7}\renewcommand{\textfraction}{0.05}

\section{Barren-plateau reproduction and correctness checks}
\label{app:bp}

\begin{figure}[tbp]
\centering
\includegraphics{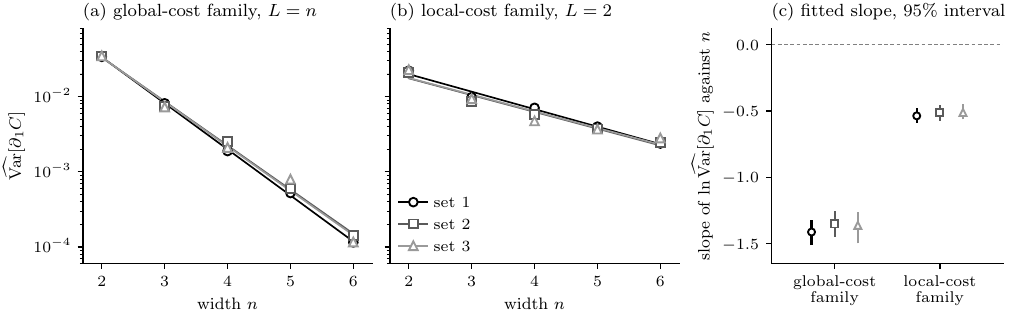}
\caption{Gradient variance against width for (a)~the global-cost family and (b)~the local-cost family
of Section~\ref{sec:setup}, with a shared vertical axis.
Here $\partial_1 C$ is the derivative of the loss $C$ with respect to the $R_Y$ angle of the first qubit
in the first layer.
In (a) and (b), each marker is the variance of $\partial_1 C$ over 200~initial points with independent,
uniformly random angles.
Each line is the least-squares fit of the natural logarithm of this variance against $n$ over one set
of random initial points, and each set draws its own points.
(c)~The fitted slope of each set for each family with its unadjusted percentile bootstrap interval
from 400 replicates that resample the derivatives at each $n$.
Marker shape and grey level identify the set, and the dashed line in (c) marks zero slope.}
\label{fig:bp}
\end{figure}

The global-cost family reproduces the exponential decay of gradient variance with width \citep{mcclean2018}, and the local-cost family decays more slowly (Section~\ref{sec:setup}). We fit the natural logarithm of the variance of $\partial_1 C$, the derivative of the loss with respect to the $R_Y$ angle of the first qubit in the first layer, against $n$ over $n = 2, \dots, 6$; at $n = 2$ the two pairs of the ring $E$ coincide, and $E$ is a single controlled-$Z$ gate. For three independent sets of random initial points, the fitted slopes of the global-cost family, with 95\% bootstrap intervals, are $-1.412$ $[-1.511, -1.321]$, $-1.351$ $[-1.446, -1.255]$ and $-1.362$ $[-1.498, -1.261]$. The local-cost family gives $-0.537$ $[-0.589, -0.480]$, $-0.514$ $[-0.574, -0.458]$ and $-0.511$ $[-0.563, -0.450]$, so its gradient variance also falls with width. Before the runs we fixed the criterion under which the local-cost family decays markedly more slowly, namely that in each set of initial points its slope has an interval containing zero or is shallower than half the global slope of the same set. Each local slope is shallower than half the global slope of its set, which is $-0.706$, $-0.676$ and $-0.681$ respectively. Figure~\ref{fig:bp} shows both families.

Parameter-shift derivatives \citep{mitarai2018}, which cost two circuit evaluations per parameter, agree with central finite differences with step $10^{-6}$ on a circuit with $n = 3$ and $L = 2$ in both cost families. For a single $R_Y$ rotation the computed metric equals $1/4$ to within $10^{-12}$, which corresponds to a quantum Fisher information of 1. The full $p \times p$ metric agrees with the metric computed from central-difference derivatives of the simulated state, with step $10^{-5}$, at five random parameter vectors for each of the eight circuits of Section~\ref{sec:setup}; over these 40 cases, whose off-diagonal entries reach 0.25 in magnitude, the largest entrywise difference is $3.3 \times 10^{-11}$. For one qubit and one layer of the global cost, the derivative with respect to the rotation angle $\theta_Y$ is $\tfrac{1}{2}\sin\theta_Y$, whose variance over a uniformly random angle is exactly $1/8$; the estimate from 2000 random angles is $0.1237$. The metric's numerical rank respects the bound $\operatorname{rank}(g) \le 2(2^n - 1)$, the real dimension of the manifold of pure $n$-qubit states, and we verify numerically that this inequality holds at $n = 2$ and $L = 4$. Like the reproduction, these checks test the simulator, with the rank check limited to the bound and no closed-form check of the Adam or SPSA update.

\section{Recomputing the conventions}
\label{app:conv}

We define the statistics of Section~\ref{sec:setup} here so that every number in Table~\ref{tab:main} can be recomputed from a table of runs.

Let a run be a record of its method, cost family, $n$, learning rate $\eta$, QNG's damping $\lambda$ where it applies, and initialization, together with its budget $B = 54p$ and the loss recorded after each step against the cumulative circuit evaluations spent. For a target $\tau$, the run's time to target $T$ is the first recorded evaluation count at which the loss is at most $\tau$. The time is $T = 0$ when the initial loss is already at most $\tau$ and $T = \varnothing$ when no recorded loss within $B$ is at most $\tau$, and a crossing at exactly $B$ counts as reached. Table~\ref{tab:app-protocol} lists the sets of runs.

\begin{table}[t]
\centering
\caption{The sets of runs of Section~\ref{sec:setup}. Initialization counts are per configuration, and the three methods share the initializations; every run is charged the per-step prices of Section~\ref{sec:setup} within its budget.}
\label{tab:app-protocol}
\small\setlength{\tabcolsep}{4pt}
\begin{tabular}{@{}p{0.95in}p{1.9in}p{1.0in}p{1.0in}p{1.45in}@{}}
\toprule
\raggedright Run set & \raggedright Settings & \raggedright Initializations & \raggedright Targets & \raggedright Used for \tabularnewline
\midrule
\raggedright Original grid & \raggedright Adam and SPSA $\eta \in \{0.1, 0.3\}$; QNG also $\lambda \in \{10^{-3}, 10^{-1}\}$ & \raggedright 10 original & \raggedright six, from 0.5 to 0.01 & \raggedright Section~\ref{sec:audit}; best original setting; fixing the strict target \tabularnewline
\addlinespace
\raggedright Extended grid & \raggedright six $\eta$ from 0.03 to 10 for Adam and SPSA; 27 settings for QNG, $\lambda$ from $10^{-5}$ to 1 & \raggedright 10 original & \raggedright 0.20, 0.01 & \raggedright The 48 selections (Table~\ref{tab:app-grid}) \tabularnewline
\addlinespace
\raggedright New initializations & \raggedright every original and every selected setting & \raggedright 20 new, held out from selection & \raggedright 0.20, 0.01; six at the strict-target selection & \raggedright Replication of Section~\ref{sec:audit}; main comparison and out-of-sample tests of Section~\ref{sec:target} \tabularnewline
\addlinespace
\raggedright Five SPSA sequences & \raggedright every SPSA setting in the rows above & \raggedright 10 original; 20 new & \raggedright 0.20, 0.01 & \raggedright Per-initialization median of five sequences (Appendix~\ref{app:spsa}) \tabularnewline
\addlinespace
\raggedright Budgets $2B$ and $4B$ & \raggedright strict-target selections; global-cost loose-target selections & \raggedright 10 original and 20 new; 10 original & \raggedright 0.20, 0.01 and final loss; final loss & \raggedright Counts and final loss at longer budgets (Appendices~\ref{app:budget} and~\ref{app:pool}) \tabularnewline
\bottomrule
\end{tabular}
\end{table}

\textbf{Reach rate} is the fraction of the $N$ runs of a group with $T \neq \varnothing$. It needs an explicit test for $\varnothing$, because $T = 0$ is a valid value (Appendix~\ref{app:zero}).

\textbf{Conditional median} is $\operatorname{median}\{T : T \neq \varnothing\}$, the summary that success-only timing reports.

\textbf{Budget-capped median} is $\operatorname{median}\{\min(T, B)\}$ over all $N$ runs, with $\min(\varnothing, B) = B$. The charge $B$ is a lower bound on the evaluations such a run would need. Throughout, the median of an even number of values is the midpoint of the two central values.

Within a configuration the budget-capped median is a Kaplan--Meier median whenever more than half of the runs reach the target. Every miss in a configuration is censored at the same $B$, so for $t < B$ the Kaplan--Meier estimate $\hat S(t)$ \citep{kaplan1958} of the probability that $T > t$ is the fraction of runs with $T > t$. Its median is taken under two conventions, the lower endpoint $\inf\{t : \hat S(t) \le 1/2\}$ and the midpoint of the interval on which $\hat S(t) = 1/2$, which reduces to the lower endpoint when $\hat S$ steps past $1/2$. With $k$ of the $N$ runs reaching the target and $N$ even, three cases arise. If $k \ge N/2 + 1$, the two central values of $\min(T, B)$ are crossing times, and the budget-capped median equals the midpoint Kaplan--Meier median. If $k = N/2$, the budget-capped median is the midpoint of the last crossing and $B$, a property of the budget rather than of the method, while the lower-endpoint median is that last crossing and the midpoint median is undefined; QNG's entry with its settings pooled at global-cost $n = 5$ in Table~\ref{tab:app-cells}, $2633.5 = (2567 + 2700)/2$, is such a case. If $k < N/2$, the Kaplan--Meier median is undefined and the budget-capped median equals $B$. The tables therefore mark every entry from a group in which at most half of the runs reach the target, since such an entry gives no time estimate.

A median pooled over configurations mixes budgets of 648 to 3888 evaluations and is not a Kaplan--Meier median. A Kaplan--Meier median of the pooled runs does not remove this problem. Each run is censored at its configuration's budget, and the configuration also sets how hard the target is, so the censoring is informative and the pooled Kaplan--Meier estimate is biased as well. At the loose target that estimate is 1168 evaluations for QNG's 320 original-grid runs, against the budget-capped median of 935 in Table~\ref{tab:main}.

The restricted mean time to target of runs $i = 1, \dots, N$ with times $T_i$ and budgets $B_i$ is $\mathrm{RMST} = N^{-1} \sum_{i=1}^{N} \min(T_i, B_i)/B_i$, the mean fraction of its budget that a run spends before it reaches the target, a miss counting as the whole budget. Its identity with the restricted mean survival time \citep[eq.~(1)]{royston2013}, the Kaplan--Meier area and the PAR-1 score, the penalized average runtime with penalty factor 1 (Section~\ref{sec:setup}), holds only within a configuration, where every miss is censored at the same budget. Because $\min(T, B)/B = \int_0^1 \mathbf{1}\{T > fB\}\,df$, with a miss counted as $T > B$, RMST is the area above the attainment curve, the share of runs with $T \le fB$ for $f \in [0, 1]$, whose value at $f = 1$ is the reach rate (Figure~\ref{fig:app-attain}). Each run is divided by its own budget, so an RMST pooled over configurations averages budget fractions and does not mix evaluation scales.

Paired comparisons match two methods at named settings through their shared initialization and use the definitions of first and tied in Section~\ref{sec:setup}, under which a pair in which both runs miss is tied. Counts are written QNG first--Adam first--tied. In Figure~\ref{fig:conventions}(b), the values of $\min(T, B)/B$ of reached runs lie on multiples of $1/27$ for Adam and of $(3p + 1)/(54p) \approx 1/18$ for QNG, so an initialization from which Adam needs $3k$ steps and QNG $2k$ steps, for a positive integer $k$, lies a factor $1 + 1/(3p)$ above the diagonal and counts as Adam first, because QNG then spends $2k$ evaluations more. The exact two-sided sign test uses the non-tied pairs. The Wilcoxon signed-rank test, the Hodges--Lehmann estimate and the median paired difference use the differences of $\min(T, B)$ in raw evaluations, so pooled over configurations they weigh configurations with larger budgets more. A comparison pooled over several settings gives one initialization several dependent pairs, and p-values are therefore quoted only for comparisons with one pair per initialization. The two exact sign tests of Section~\ref{sec:target} at the fixed strict-target selection, QNG against Adam on the new global-cost initializations at the loose and at the strict target, carry the main claim; a Bonferroni correction for these two would double their p-values, which leaves both below $10^{-9}$. Every other p-value is a per-comparison value, not adjusted for the number of tests. Final losses $F$ after the full budget are compared in decades, as differences of $\log_{10}(\max(F, 0) + 10^{-16})$.

Apart from the slopes of Appendix~\ref{app:bp}, intervals are percentile bootstrap intervals from 10\,000 replicates, or 2000 for the break-even price of Table~\ref{tab:app-price}. The main scheme resamples initializations with replacement within each configuration and uses the same draw for every method, so that each pair stays together and both medians of a ratio are resampled together. The unstratified scheme resamples the initializations of all configurations together, and Appendix~\ref{app:cells} reports it beside the main scheme where the two place an interval on different sides of 1. A statistic that depends on a selection, such as each method's best original setting, is recomputed with the selection repeated inside every replicate. Ratio intervals within a family of $k$ ratios are Bonferroni-adjusted to the level $1 - 0.05/k$. The adjusted families are two per initialization set, one at each target. Each holds ten ratios of the pooled original grid, five comparisons under both readings, namely QNG against Adam, SPSA against Adam under its shared sequence and under the per-initialization median of five sequences, and SPSA against QNG under both sequences; at the loose target the six ratios against Adam are those that Figure~\ref{fig:conventions}(a) shows. With $k = 10$ the adjusted intervals are 99.5\% intervals, each running from the 0.25th to the 99.75th percentile for nominal family-wise coverage of at least 95\%. One further family, with $k = 4$ and 98.75\% intervals, holds the factors by which success-only timing shrinks the pooled loose-target ratio to Adam, for SPSA under its shared sequence and for QNG, each on both initialization sets. Each factor is the budget-capped ratio divided by the conditional ratio on the same replicate. Every other ratio interval is an unadjusted 95\% interval and is labeled so where it appears, and intervals for paired differences and for RMST differences are unadjusted 95\% intervals.

To reproduce Table~\ref{tab:main}, take the 640 runs of the original grid with SPSA under its shared perturbation sequence, compute $T$ for each run at $\tau = 0.20$ and at $\tau = 0.01$, and pool each method's runs over its settings and the eight configurations, 160 for Adam, 320 for QNG and 160 for SPSA. Each row then reports the number of runs with $T \neq \varnothing$, the conditional median and the budget-capped median, every miss charged the budget of its own configuration.

\section{Runs that start below the loose target}
\label{app:zero}

Sixteen runs in the local-cost configuration at three qubits record a crossing at zero evaluations, and every statistic keeps them. They are four Adam, eight QNG and four SPSA runs of the original grid, from two initializations whose loss was already below the loose target, at 0.0203 and 0.1783. Such runs count as successes under the definition of time to target but carry no information about optimizer speed, and they fall in the one configuration where the Adam--SPSA order is contested (Appendix~\ref{app:cells}). We note them because a reach rate computed with a test that treats zero as missing would reclassify all sixteen as failures, lower Adam's reach at the loose target below 1, and remove the basis of the statement in Section~\ref{sec:setup} that Adam reaches the loose target in every original-grid run. On the original grid no run starts at or below the strict target. On the new initializations one initialization of the same configuration also starts at or below the loose target, and in a paired comparison at that target such an initialization gives a tie, both runs crossing at zero evaluations (Table~\ref{tab:app-budget}).

\section{Grids, selection and optimizer details}
\label{app:grid}

The grid and all 48 selections of Section~\ref{sec:setup} follow stated rules applied on the original initializations only. The extended grid contains every original setting, and its runs at those settings reproduce the 640 original-grid runs. Table~\ref{tab:app-grid} lists the selection for each method, configuration and target, and its note gives the four steps of the grid. For the best original setting of Section~\ref{sec:setup}, a tie at $B$ goes to the setting with more runs that reach the target, and the settings stay tied where no setting reaches the strict target (SPSA at global-cost $n = 6$, QNG at local-cost $n = 6$).

\begin{table}[t]
\centering
\caption{Selected learning rate $\eta$, and for QNG damping $\lambda$, for each method, configuration and target on the extended grid. The selected setting has the lowest budget-capped median over its 10 runs at the budget $B$, and ties go to the lower median final loss, then to the smaller $\eta$, then to the smaller $\lambda$. SPSA uses its shared perturbation sequence, and QNG's metric is charged at its simulator price. Section~\ref{sec:setup} defines the budget-capped median and the per-step prices. Selection used only the 10 original initializations of each configuration and none of the 20 new ones. The marks identify the 8 selections that the tie-break decided.}
\label{tab:app-grid}
\begin{tabular*}{\textwidth}{@{\extracolsep{\fill}}llcccccccc@{}}
\toprule
 & & \multicolumn{2}{c}{Adam $\eta$} & \multicolumn{4}{c}{QNG} & \multicolumn{2}{c}{SPSA $\eta$} \\
\cmidrule(lr){3-4}\cmidrule(lr){5-8}\cmidrule(l){9-10}
 & & $\tau = 0.20$ & $\tau = 0.01$ & \multicolumn{2}{c}{$\tau = 0.20$} & \multicolumn{2}{c}{$\tau = 0.01$} & $\tau = 0.20$ & $\tau = 0.01$ \\
\cmidrule(lr){5-6}\cmidrule(lr){7-8}
Family & $n$ & & & $\eta$ & $\lambda$ & $\eta$ & $\lambda$ & & \\
\midrule
global-cost & 3 & 0.3 & 0.3 & 1 & $10^{-1}$\rlap{\,$\dagger$} & 1 & $10^{-1}$ & 1 & 1 \\
 & 4 & 0.3 & 0.3 & 1 & $10^{-1}$\rlap{\,$\dagger$} & 1 & $10^{-1}$ & 1 & 1 \\
 & 5 & 0.3 & 0.3 & 1 & $10^{-2}$ & 1 & $10^{-1}$ & 1 & 1 \\
 & 6 & 0.3 & 0.3 & 1 & $10^{-2}$ & 1 & $10^{-1}$ & 1 & 1 \\
\midrule
local-cost & 3 & 1 & 0.3 & 3 & $10^{-1}$ & 1 & $10^{-3}$\rlap{\,$\ddagger$} & 3 & 3 \\
 & 4 & 1 & 0.3 & 3 & $10^{-5}$\rlap{\,$\dagger$} & 3 & $10^{-1}$ & 1 & 3 \\
 & 5 & 1 & 0.3\rlap{\,$\dagger$} & 3 & $10^{-5}$\rlap{\,$\ddagger$} & 3 & $10^{-4}$\rlap{\,$\ddagger$} & 3 & 3 \\
 & 6 & 1 & 0.3 & 3 & $10^{-2}$ & 3 & $10^{-4}$\rlap{\,$\ddagger$} & 3 & 3 \\
\bottomrule
\end{tabular*}
\par\medskip
\begin{minipage}{\textwidth}\footnotesize\relpenalty=10000
$\dagger$\,Tied with the next-ranked setting on the budget-capped median and decided by the lower median final loss. $\ddagger$\,Tied likewise, with both median final losses below $10^{-12}$; at local-cost $n = 5$, $\tau = 0.20$, both are 0 and the smaller $\lambda$ decides.

\smallskip
The extended grid was reached in four steps, each followed by a selection on the original initializations. Grid (1) has $\eta \in \{0.03, 0.1, 0.3, 1\}$ for every method and QNG $\lambda \in \{10^{-3}, 10^{-2}, 10^{-1}\}$, 160 settings over the eight configurations, and contains the original grid of Section~\ref{sec:setup}; its selection used $\tau = 0.20$ only. Grid (2) adds $\eta \in \{3, 10\}$, 240 settings; its selection and every later one used both targets. Grid (3) adds QNG $\lambda \in \{10^{-4}, 1\}$ at $\eta \in \{0.3, 1, 3\}$, 288 settings. Grid (4), the extended grid, adds QNG $\lambda = 10^{-5}$ at $\eta \in \{0.3, 1, 3\}$, 312 settings, which per configuration are 6 for Adam, 6 for SPSA and 27 for QNG. Each extension was made after selections at an edge of the previous grid. On grid (1), $\eta$ lay at its upper edge in 20 of 24 selections; on grid (2), $\lambda$ lay at an edge in 13 of 16 QNG selections and every $\eta$ lay inside; on grid (3), $\lambda$ lay at $10^{-4}$ in 4 of 16. The design allowed at most one extension of $\lambda$ beyond grid (3). On grid (4) every selected $\eta$ lies strictly inside the grid, and $\lambda$ lies inside it in 14 of the 16 QNG selections; the other two, at $\lambda = 10^{-5}$, were decided by the tie-break against $\lambda = 10^{-4}$. The selection rule applied to grid (1) at both targets selects a setting with $\eta = 1$ for the 14 local-cost entries with $\eta = 3$, and the setting shown for every other entry, all 24 global-cost entries included.
\end{minipage}
\end{table}%

On the global-cost family Adam selects $\eta = 0.3$ at both targets, which is also one of its two settings on the original grid. QNG's selections for the two targets differ only in the damping at $n = 5$ and $6$, $10^{-2}$ for the loose target against $10^{-1}$ for the strict one, and this difference underlies the dependence on the selection target in Section~\ref{sec:target}. Every global-cost selection is the same on grid~(1) of the note as on the extended grid, so the settings that Section~\ref{sec:target} holds fixed do not depend on how the grid was extended.

The tie-break decides some selections but no reported comparison. Five of the tied alternatives also have runs on the new initializations, and substituting any of them changes no per-configuration winner, no direction of a paired comparison and no per-configuration sign test across the 0.05 level.

No run at an original or a selected setting diverges, that is, ends with a loss above its initial loss. Divergence occurs at large learning rates and, for QNG, at small damping. On the original initializations 39 of the 480 Adam runs of the extended grid diverge, all at $\eta \in \{3, 10\}$, and 152 of the 2160 QNG runs diverge, at $\eta = 1$ with $\lambda \le 10^{-4}$, at $\eta = 3$ with $\lambda \le 10^{-1}$ and at $\eta = 10$ with any $\lambda$. No SPSA run and no run on the new initializations diverges.

Selecting settings on the original initializations makes them look only slightly better than they perform on the new initializations. For each setting we divide its budget-capped median on the new initializations by that on the original ones, excluding settings whose median equals the budget on both sets, and compare the geometric mean of this ratio over the selected settings with its mean over the original settings that were not selected. Over all 48 selections the two means are 1.05 and 1.04; 24 selections do worse on the new initializations, 16 do better and 8 are unchanged, and the largest change is a factor of 1.5. At the loose target the selected and unselected means are 1.077 and 1.001 for Adam, 1.037 and 1.048 for QNG and 1.163 and 1.079 for SPSA. At the strict target they are 1.019 and 1.038 for Adam, 0.978 and 1.008 for QNG and 1.016 and 1.011 for SPSA. No per-configuration winner differs between the two initialization sets, and QNG's strict-target selections do better on average on the new initializations than on the original ones.

The three optimizers use fixed constants apart from the learning rate and QNG's damping. Adam uses $\beta_1 = 0.9$, $\beta_2 = 0.999$ and $\epsilon = 10^{-8}$ with bias correction \citep{kingma2015}. QNG solves $(g + \lambda I)\,\delta = \nabla C$ exactly at every step and moves by $-\eta\,\delta$. SPSA \citep{spall1992} estimates the gradient from the loss at $\theta \pm c_t \Delta_t$, where $\Delta_t$ is a vector of independent random signs, and steps with the gain $a_t = \eta/(1 + t/10)^{0.602}$ and the perturbation size $c_t = 0.5/(1 + t/10)^{0.101}$ at step $t$, so its learning rate $\eta$ scales the step gain. The exponents 0.602 and 0.101 are the practical values of \citet[Sec.~III.A]{spall1998}. Spall's perturbation size $c_k = c/(k + 1)^{\gamma}$ has $k + 1$ where $c_t$ has $1 + t/10$, and this factor $1/10$ on the step index is our choice; $c_t$ is the same in every SPSA run and was not searched.

A rule fixed before plotting chose the pair of Figure~\ref{fig:pair}(b) as the new global-cost initialization at $n = 6$ whose strict-target difference between QNG's and Adam's $\min(T, B)$ is nearest the median, with ties going to the loose-target difference nearest its median and then to the first such initialization in the order in which the initializations were drawn. In that pair Adam reaches the loose target 655 evaluations before QNG, and QNG reaches the strict target while Adam misses it.

\section{SPSA's randomness}
\label{app:spsa}

SPSA's success-only order against Adam moves across its own perturbation sequences, while its budget-capped ratio stays above two in every sequence (Sections~\ref{sec:setup} to~\ref{sec:target}). SPSA is the only randomized method here, since Adam and QNG are deterministic given the initialization. In the original-grid runs, and in every run on the extended grid and on the new initializations, SPSA's random directions depend on the configuration, the learning rate and the evaluation count but not on the initialization, so all initializations of a configuration and learning rate share one perturbation sequence. Those runs therefore replicate over initializations under one draw of SPSA's randomness. Five independent sequences, drawn separately for each initialization, vary that randomness at every SPSA setting that was run, and Table~\ref{tab:spsa} compares them with the shared sequence on the original grid.

\begin{table}[t]
\centering
\caption{SPSA under its shared perturbation sequence and five independent sequences, on the original grid with learning rates $\eta \in \{0.1, 0.3\}$ for Adam and SPSA. Sequences 1--5 are drawn per initialization. Each entry covers 160 runs on the original initializations or 320 on the new ones (10 or 20 per configuration and learning rate), pooled over configurations with different budgets (the eight of both cost families, 648 to 3888 evaluations), so these medians mix budgets and need not equal Kaplan--Meier medians. At $\tau = 0.20$ the entries are SPSA/Adam ratios of the conditional and the budget-capped median; at $\tau = 0.01$ they are SPSA's reached runs and its conditional median in circuit evaluations, a point estimate over reached runs only. In the row for the per-initialization median of five sequences, the 160 or 320 are units; a unit is one initialization at one learning rate and reaches the target when at least three of its five sequences do. Adam's runs are the same in every row. All of them reach $\tau = 0.20$, and Adam's two medians are 280 on each initialization set; at $\tau = 0.01$ its conditional medians are 864 and 880, with 113 of 160 and 234 of 320 runs reaching the target. Section~\ref{sec:setup} defines both medians, the shared sequence and the per-initialization median of five sequences.}
\label{tab:spsa}
\small\setlength{\tabcolsep}{4.5pt}\setlength{\cmidrulekern}{4.5pt}
\begin{tabular*}{\linewidth}{@{\extracolsep{\fill}}lrrrr@{\hspace{17pt}}rrrr@{}}
\toprule
& \multicolumn{4}{c@{\hspace{17pt}}}{Original initializations} & \multicolumn{4}{@{}c@{}}{New initializations} \\
\cmidrule(lr{17pt}){2-5} \cmidrule(l{0pt}){6-9}
& \multicolumn{2}{c}{SPSA/Adam, $\tau = 0.20$} & \multicolumn{2}{c@{\hspace{17pt}}}{SPSA, $\tau = 0.01$} & \multicolumn{2}{@{}c}{SPSA/Adam, $\tau = 0.20$} & \multicolumn{2}{c@{}}{SPSA, $\tau = 0.01$} \\
\cmidrule(lr){2-3} \cmidrule(lr{17pt}){4-5} \cmidrule(l{0pt}r){6-7} \cmidrule(l){8-9}
\begin{tabular}[b]{@{}l@{}}Perturbation\\sequence\end{tabular} & Conditional & \begin{tabular}[b]{@{}r@{}}Budget-\\capped\end{tabular} & \begin{tabular}[b]{@{}r@{}}Reached\\of 160\end{tabular} & Conditional & Conditional & \begin{tabular}[b]{@{}r@{}}Budget-\\capped\end{tabular} & \begin{tabular}[b]{@{}r@{}}Reached\\of 320\end{tabular} & Conditional \\
\midrule
Shared sequence & 1.114 & 2.243 & 45 & 536 & 0.989 & 2.289 & 98 & 637 \\
\addlinespace
Sequence 1 & 1.114 & 2.229 & 46 & 532 & 1.014 & 2.304 & 89 & 674 \\
Sequence 2 & 0.979 & 2.318 & 43 & 544 & 0.971 & 2.239 & 95 & 694 \\
Sequence 3 & 1.004 & 2.211 & 46 & 653 & 0.936 & 2.221 & 94 & 657 \\
Sequence 4 & 0.886 & 2.204 & 46 & 591 & 1.032 & 2.204 & 90 & 650 \\
Sequence 5 & 0.957 & 2.146 & 48 & 549 & 1.139 & 2.314 & 85 & 684 \\
\addlinespace
\begin{tabular}[t]{@{}l@{}}Per-initialization\\median of five\\sequences\end{tabular} & 0.979 & 2.239 & 46 & 591 & 1.075 & 2.243 & 88 & 670 \\
\bottomrule
\end{tabular*}
\par\smallskip
\begin{minipage}{\linewidth}\footnotesize In three budget-capped entries at $\tau = 0.20$, one or both central values of SPSA's pooled sample come from runs of the local-cost family at $n = 3$ that miss the target and are charged that configuration's budget, 648 evaluations. For sequence~5 on the new initializations (2.314), both central values are such charges, so SPSA's budget-capped median is 648. For sequence~2 on the original initializations (2.318) and the shared sequence on the new ones (2.289), one central value is such a charge and the other a crossing time, at 650 and 634 evaluations respectively, so the medians are 649 and 641.\par SPSA's 16 selections on the extended grid (one per configuration and target) were made on the original initializations under the shared sequence. At these settings on the original initializations, the shared sequence's budget-capped median at the selection target lies below the range of the five sequences in~3~of~16 and above it in none. On the new initializations it lies above the range in none and below it in~1~of~16, the selection for the global-cost family at $n = 6$ and $\tau = 0.01$ (3497 against \mbox{3634--3777} evaluations). For six exchangeable values with no ties, the probability that a given one lies outside the range of the other five is 2/6.
\end{minipage}
\end{table}%

At the loose target the conditional ratio runs from 0.89 to 1.11 across the five sequences on the original initializations and from 0.94 to 1.14 on the new ones, so SPSA's conditional median is below Adam's under three of the five sequences on the original initializations and under two of five on the new ones. The budget-capped ratio stays between 2.15 and 2.32 under every sequence on both sets.

At the strict target SPSA's conditional median stays below Adam's under every sequence and on both initialization sets, although SPSA reaches that target in at most 48 of 160 runs on the original initializations and 98 of 320 on the new ones. These conditional medians are point estimates over a minority of runs (Appendix~\ref{app:cells}).

The shared sequence behaves as an ordinary draw among the six. Over all settings run with varied sequences, excluding those in which the five sequences agree, its loose-target budget-capped median lies outside the range of the five independent sequences in 11 of 34 settings on the original initializations and in 6 of 21 on the new ones, near the 2/6 expected when one of six exchangeable values is compared with the range of the other five. Selecting SPSA's learning rate with all five sequences changes 1 of its 16 selections, at local-cost $n = 4$ for the loose target from $\eta = 1$ to $\eta = 3$, and no global-cost selection.

With settings selected for each target, SPSA has the lowest budget-capped median of the three methods at the loose target in all eight configurations on both initialization sets, on the new initializations under each of the five sequences, and at the strict target in five of the eight configurations on the new initializations. Under the loose-target selection every Adam and SPSA run on the new initializations reaches $\tau = 0.20$, so both readings put SPSA ahead of Adam, and this direction holds at the loose target only. Taking SPSA's outcome from each new initialization as the per-initialization median of five sequences, SPSA reaches the loose target before Adam at its loose-target selection from 155 of 158 non-tied initializations, and a single sequence gives 150 to 154 of 160. Pooled over a family on the new initializations, SPSA's budget-capped median under its shared sequence is 0.365 times Adam's on the global-cost family (unadjusted 95\% interval 0.285--0.438) and 0.287 times on the local-cost family (unadjusted 95\% interval 0.200--0.333), and these intervals hold SPSA's sequence fixed. With settings selected for the strict target, the lowest budget-capped median at that target belongs to QNG at global-cost $n = 4$, $5$ and $6$ and to SPSA in the other five configurations, under every sequence. At the loose target QNG has the lowest budget-capped median in no configuration, under any grid and on either initialization set.

Against SPSA at the strict target on the global-cost family, QNG's paired verdict depends on the test. With each method's settings selected for the strict target and SPSA's outcome from each new initialization taken as the per-initialization median of five sequences, QNG reaches $\tau = 0.01$ first from 49 of the 80 global-cost initializations; the sign test gives $p = 0.057$ and the Wilcoxon signed-rank test $p = 3.5\times10^{-5}$. A single sequence gives QNG first from 47 to 54, and the shared sequence from 46, with sign-test $p = 0.22$ and Wilcoxon $p = 4.8\times10^{-4}$. The Wilcoxon test ranks raw differences, so configurations with larger budgets weigh more, and QNG's lead lies in those configurations. Under the shared sequence QNG is first from 1 of the 20 initializations at $n = 3$, 11 at $n = 4$, 14 at $n = 5$ and 20 at $n = 6$.

\section{Per-configuration results on the original grid}
\label{app:cells}

Table~\ref{tab:main} gives the pooled counts and medians of the 640-run grid behind Section~\ref{sec:audit}. We also give the runs that success-only timing drops, the conditional medians at the strict target and the medians of each configuration, which underlie the original-grid comparison of Section~\ref{sec:target}.

\begin{table*}[t]
\centering
\caption{Reach and both medians at $\tau = 0.20$ and $\tau = 0.01$ on the original grid and initializations.
Each row pools all settings of one method over eight configurations with different budgets, so its medians mix budgets and need not equal Kaplan--Meier medians.
QNG's two damping values double its number of runs; SPSA uses its shared perturbation sequence.
The Reached columns count runs that reach $\tau$.
Medians are in circuit evaluations, with the metric charged its simulator price; Section~\ref{sec:setup} defines them, the price and the budget $B$.
A superscript~${\geq}B$ marks a budget-capped median at a target that at most half the method's runs reach; both marked values equal the budget of the local-cost configuration at $n = 5$ and are not times to target.
No entry carries an interval; Figure~\ref{fig:conventions}(a) gives adjusted intervals for these medians' ratios to Adam's at $\tau = 0.20$ only.}
\label{tab:main}
\begin{tabular}{@{}lrrrrrrr@{}}
\toprule
 & & \multicolumn{3}{c}{$\tau = 0.20$} & \multicolumn{3}{c}{$\tau = 0.01$} \\
\cmidrule(lr){3-5} \cmidrule(l){6-8}
 & & & \multicolumn{2}{c}{Median} & & \multicolumn{2}{c}{Median} \\
\cmidrule(lr){4-5} \cmidrule(l){7-8}
Method & Runs & Reached & Conditional & \begin{tabular}[b]{@{}r@{}}Budget-\\capped\end{tabular} & Reached & Conditional & \begin{tabular}[b]{@{}r@{}}Budget-\\capped\end{tabular} \\
\midrule
Adam & 160 & 160 & 280 & 280 & 113 & 864 & 1000\phantom{$^{\geq B}$} \\
QNG & 320 & 188 & 584 & 935 & 85 & 776 & 1080$^{\geq B}$ \\
SPSA (shared sequence) & 160 & 109 & 312 & 628 & 45 & 536 & 1080$^{\geq B}$ \\
\bottomrule
\end{tabular}
\end{table*}%

Section~\ref{sec:audit} compares these medians with Adam's at the loose target, and Figure~\ref{fig:conventions}(a) plots the ratios with their adjusted intervals on both initialization sets. On the original initializations QNG's pooled median is 2.09 times Adam's over reached runs only and 3.34 times with misses charged the budget. On the new initializations, under the same grid, QNG's two ratios are 2.16 and 3.29, and SPSA's, under its shared sequence, are 0.99 (0.76--1.27) and 2.29 (1.78--2.68), with intervals adjusted as in Appendix~\ref{app:conv}. Success-only timing therefore shrinks the pooled ratio to Adam by a factor of 2.01 (1.46--2.62) for SPSA and 1.60 (1.37--1.87) for QNG on the original initializations, and by 2.31 (1.78--2.92) and 1.52 (1.33--1.67) on the new ones. These intervals are adjusted for the family of four factors of Appendix~\ref{app:conv}. Pooled over its four settings, QNG misses the loose target in 41.3\% of its original-grid runs on the original initializations, and SPSA, under its shared sequence, misses it in 31.9\%. Restricted to $\lambda = 10^{-3}$, QNG misses in 30.0\% of its runs, below SPSA's rate.

With learning rates pooled on the original grid and initializations at the loose target, Adam has the lowest budget-capped median in every configuration. Table~\ref{tab:app-cells} gives these medians in its left block and the medians at each method's best original setting in its right blocks. Paired with QNG at a shared learning rate on the original initializations, with both cost families pooled, Adam reaches the loose target first in 78 of the 78 non-tied pairs at each of QNG's four settings.

\begin{table}[t]
\centering
\caption{Budget-capped median circuit evaluations to the target per configuration, on the original grid and
initializations, with SPSA under its shared perturbation sequence and QNG's metric at its simulator price.
Section~\ref{sec:setup} and Appendix~\ref{app:conv} define this median and relate it to the Kaplan--Meier median.
${=}B$ marks an entry equal to the budget $B$, and ${\geq}B$ any other entry from a group in which at most half of
the runs reach the target. A marked entry gives no time estimate. The left block is at $\tau = 0.20$ with each
method's settings pooled, 20~runs per entry and 40 for QNG. The right blocks take each method's best original
setting (Section~\ref{sec:setup}), 10 runs per entry.}
\label{tab:app-cells}
\small\setlength{\tabcolsep}{7.5pt}
\begin{tabular}{@{}lrrr@{}lr@{}l@{\hspace{29pt}}r@{}l@{\hspace{36pt}}r@{}lr@{}lr@{}l@{\hspace{22pt}}r@{}lr@{}l@{\hspace{29pt}}r@{}l@{\hspace{14pt}}}
\toprule
 & & & \multicolumn{6}{c@{\hspace{36pt}}}{$\tau = 0.20$, settings pooled} & \multicolumn{6}{@{}c@{\hspace{22pt}}}{$\tau = 0.20$, best setting} & \multicolumn{6}{@{}c@{\hspace{14pt}}}{$\tau = 0.01$, best setting} \\
\cmidrule(l{7.5pt}r{36pt}){4-9}\cmidrule(l{0pt}r{22pt}){10-15}\cmidrule(l{0pt}r{14pt}){16-21}
Family & $n$ & $B$ & \multicolumn{2}{c}{\makebox[0pt]{Adam}} & \multicolumn{2}{c@{\hspace{29pt}}}{\makebox[0pt]{QNG}} & \multicolumn{2}{@{}c@{\hspace{36pt}}}{\makebox[0pt]{SPSA}} & \multicolumn{2}{@{}c}{\makebox[0pt]{Adam}} & \multicolumn{2}{c}{\makebox[0pt]{QNG}} & \multicolumn{2}{c@{\hspace{22pt}}}{\makebox[0pt]{SPSA}} & \multicolumn{2}{@{}c}{\makebox[0pt]{Adam}} & \multicolumn{2}{c@{\hspace{29pt}}}{\makebox[0pt]{QNG}} & \multicolumn{2}{@{}c@{\hspace{14pt}}}{\makebox[0pt]{SPSA}} \\
\midrule
global-cost & 3 & 972 & 198 &  & 522 & .5 & 326 &  & 144 &  & 275 &  & 127 &  & 702 &  & 385 &  & 304 &  \\
 & 4 & 1728 & 320 &  & 970 &  & 848 &  & 288 &  & 533 & .5 & 338 &  & 1504 &  & 776 &  & 1009 &  \\
 & 5 & 2700 & 650 &  & 2633 & .5\rlap{\smash{$^{\geq B}$}} & 2700 & \rlap{\smash{$^{=B}$}} & 450 &  & 1434 & .5 & 844 &  & 2400 &  & 1812 &  & 2700 & \rlap{\smash{$^{=B}$}} \\
 & 6 & 3888 & 1008 &  & 3888 & \rlap{\smash{$^{=B}$}} & 3888 & \rlap{\smash{$^{=B}$}} & 720 &  & 2821 &  & 3505 &  & 3672 &  & 3255 &  & 3888 & \rlap{\smash{$^{=B}$}} \\
\addlinespace
local-cost & 3 & 648 & 84 &  & 222 &  & 98 &  & 60 &  & 129 & .5 & 28 &  & 312 &  & 370 &  & 302 &  \\
 & 4 & 864 & 128 &  & 661 & .5 & 287 &  & 96 &  & 318 & .5 & 117 &  & 576 &  & 864 & \rlap{\smash{$^{=B}$}} & 586 &  \\
 & 5 & 1080 & 260 &  & 1080 & \rlap{\smash{$^{=B}$}} & 728 &  & 140 &  & 640 & .5 & 315 &  & 820 &  & 1080 & \rlap{\smash{$^{=B}$}} & 1080 & \rlap{\smash{$^{=B}$}} \\
 & 6 & 1296 & 168 &  & 1168 &  & 350 &  & 120 &  & 657 &  & 176 &  & 912 &  & 1296 & \rlap{\smash{$^{=B}$}} & 1191 & \rlap{\smash{$^{\geq B}$}} \\
\bottomrule
\end{tabular}
\end{table}%

In the pooled block the lower of QNG's and SPSA's entries is 1.17 (local-cost $n = 3$) to 4.05 (global-cost $n = 5$) times Adam's, and 4.05 rests on a ${\geq}B$ entry. At local-cost $n = 3$ the margin of 1.17 between SPSA's entry and Adam's is too thin to interpret as a separation, and this is the configuration whose order success-only timing reverses. By contrast, with learning rates selected on the original initializations and tested on new ones, Adam's budget-capped median is the lowest in no configuration at either target.

Table~\ref{tab:app-gap} gives SPSA's ratio to Adam under both readings in every configuration and on both initialization sets, the per-configuration counterpart of the pooled ratios of Section~\ref{sec:audit}. The reading over reached runs only shrinks SPSA's ratio within a configuration by factors from 1.04 (global-cost $n = 3$, new initializations) to 3.62 (global-cost $n = 5$, original initializations), and the top of this range rests on an entry in which SPSA's budget-capped median equals $B$ (marked ${=}B$) and so gives no time estimate. The pooled conditional median hides SPSA's gap to Adam because SPSA's misses fall unevenly across configurations. On the original initializations 25 of SPSA's 51 misses fall in the global-cost configurations at $n = 5$ and $6$, which hold a quarter of the runs, so these two configurations supply 15 of SPSA's 109 reached runs against 40 of Adam's 160. These two configurations have the largest budgets and the longest times, and SPSA's pooled conditional median therefore draws on configurations with smaller budgets and shorter times than the configurations behind Adam's pooled median. Dividing each time by its configuration's budget before pooling removes this difference in scale, and success-only timing still about halves SPSA's pooled ratio to Adam, from 2.66 to 1.32 on the original and from 2.34 to 1.22 on the new initializations. The pooled conditional ratio lies below the per-configuration conditional ratio in 7 of the 8 configurations, and on the new initializations, where the same two configurations hold 42 of SPSA's 104 misses, in 6 of the 8. Under SPSA's shared sequence the conditional ratio is at most 1 while the budget-capped ratio exceeds 1 in one configuration on the original initializations, local-cost $n = 3$, and in two on the new ones, where global-cost $n = 3$ joins it with a budget-capped ratio within 1\% of parity. Under SPSA's per-initialization median of five sequences the same holds in one configuration on the original initializations and in three on the new ones, global-cost $n = 3$ and $4$ and local-cost $n = 3$.

\begin{table}[t]
\centering
\caption{SPSA against Adam in each configuration on the original grid at the loose target $\tau = 0.20$, with SPSA under its shared perturbation sequence and each method's two learning rates pooled, 20~runs per method and configuration on the original initializations and 40 on the new ones. Each block gives SPSA's reached runs and the ratio of SPSA's median to Adam's for the conditional median over reached runs only and the budget-capped median with misses charged the budget~$B$ (Section~\ref{sec:setup}). Adam reaches the target in all its runs, so its two medians coincide, and the two ratios differ only through SPSA's misses. A ratio marked ${=}B$ has SPSA's budget-capped median equal to~$B$ because fewer than half of SPSA's runs reach the target. Such a ratio is $B$ over Adam's median and gives no time estimate.}
\label{tab:app-gap}
\small\setlength{\tabcolsep}{7pt}
\begin{tabular}{@{}lrrrrr@{\hspace{24pt}}rrr@{\hspace{14pt}}}
\toprule
& & & \multicolumn{3}{c@{\hspace{24pt}}}{Original initializations} & \multicolumn{3}{c@{\hspace{14pt}}}{New initializations} \\
\cmidrule(lr{24pt}){4-6} \cmidrule(l{0pt}r{14pt}){7-9}
& & & SPSA & \multicolumn{2}{c@{\hspace{24pt}}}{SPSA/Adam} & \multicolumn{1}{@{}r}{SPSA} & \multicolumn{2}{c@{\hspace{14pt}}}{SPSA/Adam} \\
\cmidrule(lr{24pt}){5-6} \cmidrule(lr{14pt}){8-9}
Family & $n$ & $B$ & \begin{tabular}[b]{@{}r@{}}Reached\\of 20\end{tabular} & Conditional & \begin{tabular}[b]{@{}r@{}}Budget-\\capped\end{tabular} & \begin{tabular}[b]{@{}r@{}}Reached\\of 40\end{tabular} & Conditional & \begin{tabular}[b]{@{}r@{}}Budget-\\capped\end{tabular} \\
\midrule
global-cost & 3 & 972 & 18 & 1.348 & 1.646 & 35 & 0.972 & 1.009 \\
 & 4 & 1728 & 15 & 1.269 & 2.650 & 27 & 1.057 & 2.659 \\
 & 5 & 2700 & 9 & 1.148 & 4.154\rlap{\smash{$^{=B}$}} & 23 & 1.333 & 3.070 \\
 & 6 & 3888 & 6 & 2.245 & 3.857\rlap{\smash{$^{=B}$}} & 15 & 2.366 & 4.500\rlap{\smash{$^{=B}$}} \\
\addlinespace
local-cost & 3 & 648 & 17 & 0.381 & 1.167 & 33 & 0.958 & 1.365 \\
 & 4 & 864 & 17 & 1.250 & 2.242 & 30 & 1.312 & 2.008 \\
 & 5 & 1080 & 12 & 1.431 & 2.800 & 28 & 1.488 & 2.638 \\
 & 6 & 1296 & 15 & 1.333 & 2.083 & 25 & 1.454 & 2.505 \\
\bottomrule
\end{tabular}
\end{table}%

\begin{figure}[tbp]
\centering
\includegraphics{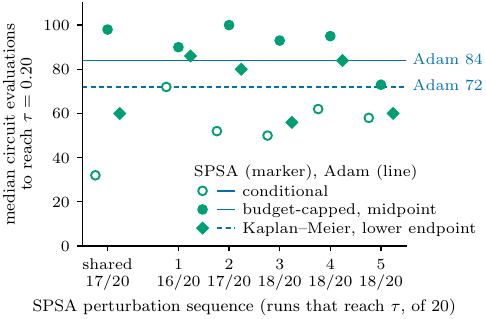}
\caption{SPSA against Adam on the original grid in the local-cost configuration at $n = 3$, at the loose target $\tau = 0.20$ with budget $B = 648$ circuit evaluations and the learning rates $\eta = 0.1$ and $0.3$ pooled. Adam and each SPSA sequence have 20~runs from the 10~original initializations (definitions in Section~\ref{sec:setup}).
SPSA uses its shared perturbation sequence or one of five independent sequences (1--5), drawn per initialization.
Open circles mark the conditional median, filled circles the budget-capped median and diamonds the lower-endpoint Kaplan--Meier median, here the lower of the two central values of $\min(T, B)$ (Appendix~\ref{app:conv}).
Each marker is compared with Adam's line for the same convention. Adam reaches $\tau$ in every run, so its conditional and budget-capped medians are equal and share the solid line. The dashed line is its lower-endpoint Kaplan--Meier median.
Runs from the two initializations that start below $\tau$ record a crossing at zero evaluations and are kept in every median (Appendix~\ref{app:zero}).}
\label{fig:app-l3}
\end{figure}

The reversal on the original initializations, in the local-cost configuration at $n = 3$ with learning rates pooled, rests on SPSA's perturbation sequence and on how the median of an even sample is taken. Figure~\ref{fig:app-l3} shows SPSA's three medians in this configuration under its shared sequence and under five independent sequences, each against Adam's median under the same convention.

Under the shared sequence, the three of SPSA's 20 runs that miss the loose target are what separate its conditional median of 32 evaluations from its budget-capped median of 98, against Adam's 84 under both. Once those misses are charged the budget, the two median conventions disagree about whether SPSA is behind Adam, since SPSA's lower-endpoint Kaplan--Meier median is 60 against Adam's 72. With learning rates pooled and under SPSA's shared sequence, this reversal holds under the midpoint median only. Under the five independent sequences SPSA's conditional median lies between 50 and 72, below Adam's 84 in each, so success-only timing puts SPSA ahead under every sequence. Its budget-capped median lies between 73 and 100 and exceeds Adam's 84 in four of the five, and its lower-endpoint Kaplan--Meier median lies between 56 and 86 and exceeds Adam's 72 in three; in those three both median conventions put SPSA behind Adam. With learning rates not pooled, at $\eta = 0.3$, SPSA's budget-capped median under the shared sequence is 28 against Adam's 60, the entries of Table~\ref{tab:app-cells} at the best original setting. On the new initializations the reversal recurs with learning rates pooled over the original grid, at both targets, and it does not occur at any selected setting on either initialization set.

At each method's best original setting, the order at the strict target differs between the two cost families. On the global-cost family QNG's budget-capped median at $\tau = 0.01$ is below Adam's in all four configurations and the lowest of the three methods at $n = 4$, $5$ and $6$, while SPSA's is the lowest at $n = 3$. These are the per-configuration medians behind the original-grid comparison of Section~\ref{sec:target}, in which QNG reaches the strict target first from 37 of 40 global-cost initializations (exact sign test, $p = 1.9\times10^{-8}$). On the local-cost family QNG's entries at $n = 4$, $5$ and $6$ equal the budget, since QNG reaches the strict target in at most 2 of 10 runs there, and Adam's entries are the lowest at those widths. At the loose target Adam's entry is below QNG's in all eight configurations, while SPSA's is below Adam's at $n = 3$ in both families.

Pooled over the eight configurations, the loose-target runs at each method's best original setting give QNG a budget-capped median of 565.5 evaluations against Adam's 160, the ratio of 3.53 that Section~\ref{sec:audit} reports. Resampling initializations within configurations gives an unadjusted 95\% interval of 2.70--4.04, and resampling the 80 original initializations without regard to configuration gives an unadjusted 95\% interval of 2.34--4.05, with the best original setting selected again inside every replicate. Both intervals lie above 1. At this setting QNG misses the loose target in 7.50\% of its 80 runs, SPSA under its shared sequence in 8.75\% and Adam in none. QNG's budget-capped median there is 1.148 times its conditional median, against 1.60 over the original grid. Under the loose-target selection all 80 of QNG's runs on the original initializations reach the target, and its two medians are equal.

The runs that success-only timing drops end far above the loose target when QNG's settings are pooled, but at QNG's best original setting several of them end just short of it.
On the original grid and initializations QNG reaches the loose target in half of its runs in the global-cost configuration at $n = 5$ and in a quarter at $n = 6$.
Pooled over its four settings, its runs that miss there end the budget at median losses of 0.767 and 0.944, against a target of 0.20.
At its best original setting QNG misses the target in 6 of 80 runs; three of these end within 0.015 of the target, and all six are still decreasing at their last step.
Success-only timing discards both kinds of run alike, those that end far from the target and those that the budget stops while their loss is still falling.

On the original grid and initializations at the strict target, the point estimates of the conditional medians rank the method that reaches the target most often as the slowest.
With the metric at its simulator price and SPSA under its shared sequence, Adam reaches $\tau = 0.01$ in 70.6\% of its runs, SPSA in 28.1\% and QNG in 26.6\%, yet over reached runs only Adam's median is the highest of the three, 864 evaluations against 776 for QNG and 536 for SPSA (Table~\ref{tab:main}).
In the adjusted families of Appendix~\ref{app:conv}, the intervals of the two readings lie on opposite sides of~1 only for SPSA against Adam at the strict target on the new initializations under the within-configuration scheme, where the conditional ratio is 0.724 (0.552--0.880) and the budget-capped ratio 1.111 (1.071--1.164) under SPSA's shared sequence.
In that comparison both budget-capped medians fall on a configuration's budget, and the unstratified scheme does not separate the two intervals.

\section{Attainment curves and restricted mean time to target}
\label{app:attain}

\begin{figure}[tbp]
\centering
\includegraphics{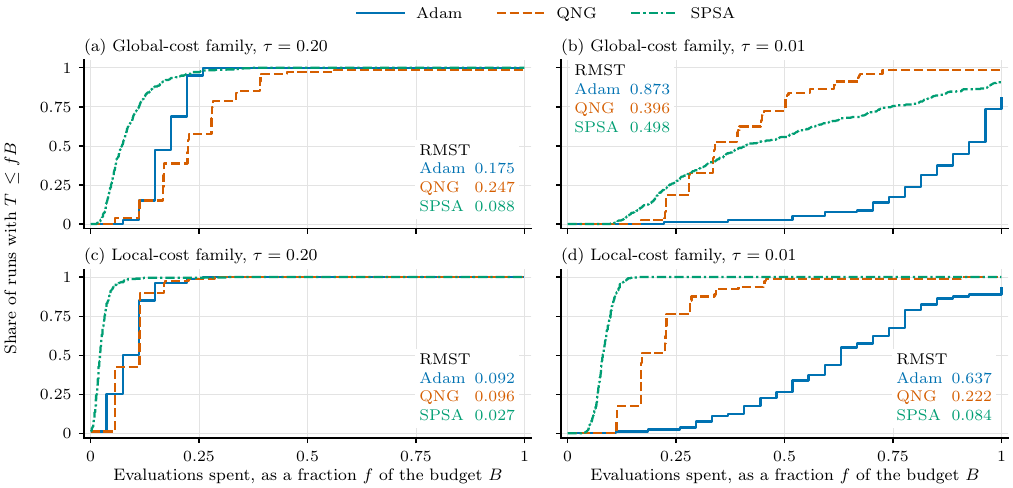}
\caption{Attainment over the budget with each target's own selection. Rows are the two cost families, each pooled over its four configurations ($n = 3$--$6$ qubits) on the 20 new initializations per configuration. Columns are the two targets; in each column every method runs, in each configuration, at the setting selected for that target on the original initializations (Table~\ref{tab:app-grid}). SPSA contributes five runs per initialization, one under each of five independent perturbation sequences. The curve at $f \in [0, 1]$ is the share of a method's runs with $T \le fB$, where $T$ and the budget $B$ of each run's configuration are those of Section~\ref{sec:setup} and QNG's metric is charged its simulator price. The curve at $f = 1$ is the method's reach rate. The value printed for each method is its restricted mean time to target (RMST), the area between its curve and 1 over $0 \le f \le 1$, with misses charged the budget.}
\label{fig:app-attain}
\end{figure}

For the comparison by restricted mean time to target (RMST) in Section~\ref{sec:target}, Figure~\ref{fig:app-attain} shows where along the budget each method reaches each target with the settings selected for that target, and the RMST printed in each panel for each method summarizes that method's curve in one number. Because each target uses its own selection, the loose-target column runs QNG with damping $10^{-2}$ at $n = 5$ and $6$ on the global-cost family (Table~\ref{tab:app-grid}), so in those configurations it differs from the settings held fixed in Figure~\ref{fig:conventions}(b).

At the strict target the curves separate early in the budget. At half the budget the share of runs that have reached $\tau = 0.01$ is 0.725 for QNG on the global-cost family and 0.988 on the local-cost family, 0.557 and 1.000 for SPSA, and 0.025 and 0.263 for Adam. Adam's curves rise later and end at 0.812 and 0.938 at the full budget, while QNG's global-cost curve ends at its reach rate of 0.988. On the global-cost family the gap between QNG's and Adam's curves is therefore far wider at half the budget than at the full budget.

The paired differences in RMST, QNG minus Adam with 95\% within-configuration bootstrap intervals, are $+0.072$ (0.050--0.099) at the loose target and $-0.476$ ($-0.515$ to $-0.435$) at the strict target on the global-cost family, and $+0.005$ ($-0.004$ to 0.014) and $-0.416$ ($-0.456$ to $-0.376$) on the local-cost family. Under the same selections, which use different runs at the two targets, Adam is first on the global-cost family from 66 of the 80 new initializations at the loose target, and QNG is first from 77 at the strict target, where the selection is the one held fixed in Section~\ref{sec:target}. The local-cost interval at the loose target contains zero (Appendix~\ref{app:select}).

\section{Selection target against evaluation target}
\label{app:select}

Table~\ref{tab:app-select} gives the paired QNG--Adam verdict for every combination of the target that selected the settings and the target that evaluates them. On the global-cost family the strict-target selection gives the reversal of Section~\ref{sec:target}, with Adam first at the loose target and QNG first at the strict target. Under that selection Adam is first at both targets from 3 of the 80 new initializations and QNG from 12, and the first method changes from Adam to QNG for the other 65, so every initialization from which QNG is first at the loose target is also one from which it is first at the strict target. The direction is the same among the global-cost pairs in which both runs reach the target, which leave out Adam's 15 misses of the strict target, with QNG first in 12 of 79 at the loose and 62 of 64 at the strict target. The loose-target selection keeps Adam first at the loose target, but at the strict target and on final loss QNG's lead over the family is no longer significant (sign test $p = 0.23$ and $0.22$).

\begin{table*}[t]
\centering
\caption{QNG against Adam on the 20 new initializations per configuration, by the target that selected the settings and the target at which they are evaluated. Settings were selected on the original initializations at the strict target $\tau = 0.01$ or the loose target $\tau = 0.20$; Adam searches six learning rates, QNG 27 settings of learning rate $\eta$ and damping $\lambda$ or, in the restricted row, Adam's six learning rates at $\lambda = 10^{-3}$. A count gives the pairs sharing an initialization in which QNG is first, Adam is first, or the two tie; at a target, first means reaching $\tau$ with fewer circuit evaluations, a miss charged the budget $B$, and on final loss it means the lower loss after the full budget. Every row compares one setting per method in each configuration, so each initialization gives one pair and every count carries its exact two-sided sign-test $p$, ties dropped. QNG's metric is charged at its simulator price; Section~\ref{sec:setup} defines the cost families, the initialization sets and the per-step prices. (a)~Each family pooled over its four configurations, $n = 3$--$6$ qubits, 80 pairs per row. (b)~The global-cost family under the loose-target selection, 20 pairs per configuration; Adam selects $\eta = 0.3$ in every global-cost configuration under both targets.}
\label{tab:app-select}
\small
\begin{tabular}{@{}lll cc cc cc@{}}
\toprule
\multicolumn{3}{@{}l}{(a) Each family pooled over $n$} & \multicolumn{2}{c}{Evaluated at $\tau = 0.20$} & \multicolumn{2}{c}{Evaluated at $\tau = 0.01$} & \multicolumn{2}{c}{Final loss} \\
\cmidrule(lr){4-5} \cmidrule(lr){6-7} \cmidrule(l){8-9}
Selected at & Family & QNG search & Count & $p$ & Count & $p$ & Count & $p$ \\
\midrule
$\tau = 0.01$ & global-cost & 27 settings & 12--68--0 & $1.2\times10^{-10}$ & 77--3--0 & $1.4\times10^{-19}$ & 79--1--0 & $1.3\times10^{-22}$ \\
 &  & 6 at $\lambda = 10^{-3}$ & 1--79--0 & $1.3\times10^{-22}$ & 73--7--0 & $5.8\times10^{-15}$ & 76--4--0 & $2.8\times10^{-18}$ \\
 & local-cost & 27 settings & 36--43--1 & $0.50$ & 79--1--0 & $1.3\times10^{-22}$ & 80--0--0 & $1.7\times10^{-24}$ \\
\midrule
$\tau = 0.20$ & global-cost & 27 settings & 14--66--0 & $3.1\times10^{-9}$ & 40--29--11 & $0.23$ & 46--34--0 & $0.22$ \\
 & local-cost & 27 settings & 24--55--1 & $6.4\times10^{-4}$ & 78--2--0 & $5.4\times10^{-21}$ & 80--0--0 & $1.7\times10^{-24}$ \\
\bottomrule
\end{tabular}
\par\medskip
\begin{tabular}{@{}lll cc cc@{}}
\toprule
\multicolumn{3}{@{}l}{(b) Global-cost family, per configuration} & \multicolumn{2}{c}{Evaluated at $\tau = 0.01$} & \multicolumn{2}{c}{Final loss} \\
\cmidrule(lr){4-5} \cmidrule(l){6-7}
Selected at & Qubits & QNG $(\eta, \lambda)$ & Count & $p$ & Count & $p$ \\
\midrule
$\tau = 0.20$ & $n = 3$ & $(1, 10^{-1})$ & 18--2--0 & $4.0\times10^{-4}$ & 20--0--0 & $1.9\times10^{-6}$ \\
 & $n = 4$ & $(1, 10^{-1})$ & 19--1--0 & $4.0\times10^{-5}$ & 19--1--0 & $4.0\times10^{-5}$ \\
 & $n = 5$ & $(1, 10^{-2})$ & 2--14--4 & $0.0042$ & 2--18--0 & $4.0\times10^{-4}$ \\
 & $n = 6$ & $(1, 10^{-2})$ & 1--12--7 & $0.0034$ & 5--15--0 & $0.041$ \\
\bottomrule
\end{tabular}
\end{table*}%

Panel~(b) locates this change in two configurations. At $n = 3$ and $4$ the two targets select the same QNG setting, $(\eta, \lambda) = (1, 10^{-1})$, and QNG reaches the strict target first from 37 of the 40 new initializations. At $n = 5$ and $6$ the loose-target selection lowers QNG's damping to $10^{-2}$. QNG then reaches the strict target from only 3 and 1 of the 20 new initializations, its budget-capped median equals the budget (2700 and 3888 evaluations, against Adam's 2600 and 3744), and Adam is first and ends at the lower final loss in most pairs. On the global-cost family the verdict at the strict target and on final loss therefore depends on the target that selected QNG's damping. \citet[App.~B]{sung2020} report a dependence of the same kind for SPSA across precision cutoffs.

The restricted row gives the counts of Section~\ref{sec:target} for QNG's search restricted to Adam's six learning rates, under which QNG also ends with the lower final loss from 76 of the 80 new global-cost initializations.

On the local-cost family the target that selected the settings changes the verdict only at the loose target. With each target's own selection the paired direction reverses as on the global-cost family, although the two targets are then evaluated on different runs. Adam is first at the loose target from 55 of 79 non-tied initializations (sign test $p = 6.4\times10^{-4}$), and in every local-cost configuration Adam is first in most non-tied pairs, with QNG first from 6 of 19, 5 of 20, 7 of 20 and 6 of 20. QNG is first at the strict target from 79 of 80. The loose-target side is weaker than on the global-cost family. The Wilcoxon signed-rank test gives $p = 0.060$, the RMST difference has an interval that contains zero (Appendix~\ref{app:attain}), and under the strict-target selection the loose-target count becomes 36--43--1 (sign test $p = 0.50$). At the strict target and on final loss QNG leads under both selections.

Table~\ref{tab:app-targets} holds the strict-target selection fixed and gives the paired counts on both families at the six examined targets of Section~\ref{sec:setup}. Every count comes from the new initializations, which were run after the strict target was fixed. On the global-cost family the 35--45 count at $\tau = 0.1$ is not significant (sign test $p = 0.31$), while the counts on either side of it are. QNG's single global-cost miss comes from one initialization at $n = 4$ and recurs at every target. With QNG's local-cost settings at $n = 5$ and $6$ taken from grid~(2) of the note to Table~\ref{tab:app-grid}, $(\eta, \lambda) = (3, 10^{-3})$ instead of $(3, 10^{-4})$, the local-cost counts change only at $\tau = 0.20$, to 38--41--1, and at $\tau = 0.1$, to 53--27--0, so the majority switches to QNG between the same two targets.

\begin{table*}[t]
\centering
\caption{QNG against Adam at six targets on the 80 new initializations of each cost family, 20 in each configuration with $n = 3$--$6$ qubits. Each method runs at the setting selected per configuration for the strict target $\tau = 0.01$ on the original initializations and held fixed at every target (the $\tau = 0.01$ columns of Table~\ref{tab:app-grid}). A count gives the pairs sharing an initialization in which QNG reaches $\tau$ with fewer circuit evaluations, in which Adam does, and in which the two tie, a miss charged the budget $B$. Every tie is a pair whose initial loss is already at or below $\tau$, and every count carries its exact two-sided sign-test~$p$, ties dropped. The rows Adam reached and QNG reached count each method's runs that reach $\tau$ within $B$. QNG's metric is charged at its simulator price; Section~\ref{sec:setup} defines the cost families, the initialization sets and the per-step prices. The columns at $\tau = 0.20$ and $0.01$ repeat the rows of Table~\ref{tab:app-select}(a) selected at $\tau = 0.01$ with QNG's 27-setting search.}
\label{tab:app-targets}
\small
\begin{tabular}{@{}l cccccc@{}}
\toprule
Target $\tau$ & 0.50 & 0.20 & 0.10 & 0.05 & 0.02 & 0.01 \\
\midrule
\multicolumn{7}{@{}l}{Global-cost family} \\
\quad QNG first--Adam first--tied & 3--77--0 & 12--68--0 & 35--45--0 & 63--17--0 & 75--5--0 & 77--3--0 \\
\quad Sign-test $p$ & $1.4\times10^{-19}$ & $1.2\times10^{-10}$ & $0.31$ & $2.3\times10^{-7}$ & $4.3\times10^{-17}$ & $1.4\times10^{-19}$ \\
\quad Adam reached & 80 & 80 & 80 & 80 & 80 & 65 \\
\quad QNG reached & 79 & 79 & 79 & 79 & 79 & 79 \\
\midrule
\multicolumn{7}{@{}l}{Local-cost family} \\
\quad QNG first--Adam first--tied & 12--25--43 & 36--43--1 & 52--28--0 & 60--20--0 & 76--4--0 & 79--1--0 \\
\quad Sign-test $p$ & $0.047$ & $0.50$ & $0.0097$ & $8.6\times10^{-6}$ & $2.8\times10^{-18}$ & $1.3\times10^{-22}$ \\
\quad Adam reached & 80 & 80 & 80 & 80 & 80 & 75 \\
\quad QNG reached & 80 & 80 & 80 & 80 & 80 & 80 \\
\bottomrule
\end{tabular}
\end{table*}%

\section{Break-even price of the metric}
\label{app:price}

We obtain the break-even price $m^\ast$ of Section~\ref{sec:target} by re-pricing the simulated runs, and Table~\ref{tab:app-price} gives it for each configuration. The hardware cost of the full metric grows quadratically with the number of parameters \citep[Sec.~3.1]{gacon2021}. For circuits in which each parameter enters through one Pauli rotation, \citet[Sec.~II.B.3]{wierichs2020} count $p(p+1)/2 + p$ circuits for the full metric, the gradient not included. The count $p(p+1)/2$ assumed here lies $p$ below theirs, so wherever the assumed count exceeds $m^\ast$, their count does too. Re-pricing charges each QNG step $2p + m$ evaluations instead of $3p + 1$ and keeps only the steps that fit within $B$. The noiseless trajectories are reused unchanged, so the price changes only how far along its trajectory a QNG run gets within the budget. A higher $m$ multiplies the evaluation count after every QNG step by ${(2p + m)/(3p + 1)}$ and can push crossings beyond $B$, so it never raises the number of initializations from which QNG reaches a target before Adam, and a high enough $m$ decreases it.

\begin{table*}[tp]
\centering
\caption{Break-even price of the metric. Each $m^\ast$ is the largest per-step price $m$ of the metric (Section~\ref{sec:setup}) at which QNG's budget-capped median at $\tau = 0.01$, or in the final-loss columns its median final loss after the last step that fits within $B$, stays below Adam's; ``none'' means that QNG's median is not below Adam's even at $m = p + 1$. Every $m^\ast$ is computed by re-pricing noiseless simulated runs and is not a hardware measurement. The simulator price is $m = p + 1$, and the assumed hardware count $p(p{+}1)/2$ charges one circuit for each independent entry of the symmetric $p \times p$ metric. Outside the column for the best original setting, Adam runs at learning rate $\eta = 0.3$ and QNG at learning rate and damping $(\eta, \lambda) = (1, 10^{-1})$ on the global-cost family and $(1, 10^{-3})$, $(3, 10^{-1})$, $(3, 10^{-4})$ and $(3, 10^{-4})$ on the local-cost family at $n = 3$--$6$; these settings were selected for $\tau = 0.01$ on the 10 original initializations per configuration and never on the 20 new ones. The best original setting of Section~\ref{sec:setup}, QNG at $(0.3, 10^{-3})$ and Adam at $\eta = 0.3$, is evaluated on the original initializations. Intervals are 95\% within-configuration percentile bootstrap intervals over 2000 replicates; a replicate in which QNG's median is not below Adam's at $m = p + 1$ counts as $m^\ast = p$, and $f$ is the fraction of replicates with $m^\ast \ge p(p{+}1)/2$.}
\label{tab:app-price}
\small
\setlength{\tabcolsep}{4pt}
\begin{tabular}{rrr r@{\ }ll r@{\ }ll c r@{\ }ll}
\toprule
& & & \multicolumn{7}{c}{Budget-capped time to $\tau = 0.01$} & \multicolumn{3}{c}{Final loss} \\
\cmidrule(lr){4-10}\cmidrule(lr){11-13}
& & & \multicolumn{3}{c}{New initializations} & \multicolumn{3}{c}{Original initializations} & \multicolumn{1}{c}{\begin{tabular}[b]{@{}c@{}}Best original\\setting\end{tabular}} & \multicolumn{3}{c}{New initializations} \\
\cmidrule(lr){4-6}\cmidrule(lr){7-9}\cmidrule(lr){10-10}\cmidrule(lr){11-13}
$n$ & $p{+}1$ & $p(p{+}1)/2$ & \multicolumn{2}{c}{$m^\ast$ (95\% interval)} & $f$ & \multicolumn{2}{c}{$m^\ast$ (95\% interval)} & $f$ & \multicolumn{1}{c}{$m^\ast$} & \multicolumn{2}{c}{$m^\ast$ (95\% interval)} & $f$ \\
\midrule
\multicolumn{13}{l}{\emph{Global-cost family ($L = n$)}} \\
3 & 19 & 171 & 161 & (91--197) & 0.13 & 104 & (97--179) & 0.06 & 64 & 158 & (126--207) & 0.14 \\
4 & 33 & 528 & 223 & (170--255) & 0 & 236 & (179--249) & 0 & 123 & 224 & (182--281) & 0 \\
5 & 51 & 1275 & 271 & (188--333) & 0 & 199 & (155--249) & 0 & 99 & 285 & (200--350) & 0 \\
6 & 73 & 2628 & 271 & (247--341) & 0 & 223 & (132--305) & 0 & 100 & 288 & (209--342) & 0 \\
\midrule
\multicolumn{13}{l}{\emph{Local-cost family ($L = 2$)}} \\
3 & 13 & 78 & 53 & (39--91) & 0.13 & 79 & (23--135) & 0.48 & none & 138 & (105--138) & 0.9995 \\
4 & 17 & 136 & 170 & (107--197) & 0.70 & 159 & (67--202) & 0.62 & none & 184 & (140--256) & 0.99 \\
5 & 21 & 210 & 219 & (149--239) & 0.56 & 194 & (111--266) & 0.29 & none & 230 & (230--320) & 0.99 \\
6 & 25 & 300 & 205 & (155--271) & 0.0005 & 255 & (157--335) & 0.11 & none & 276 & (211--384) & 0.20 \\
\bottomrule
\end{tabular}
\par\vspace{5pt}
\begin{minipage}{\textwidth}\small Paired over the 80 new initializations of each cost family at $\tau = 0.01$, the counts of pairs in which QNG is first, Adam is first or the two tie are 77--3--0 on the global-cost family and 79--1--0 on the local-cost family at $m = p + 1$, and 3--56--21 and 32--43--5 at $m = p(p{+}1)/2$; a~method is first when it reaches the target with fewer circuit evaluations, a miss charged the budget. Each pair compares the two runs from one initialization, whereas $m^\ast$ compares medians.\end{minipage}
\end{table*}%

On the global-cost family the break-even price grows far more slowly than the assumed count. With the strict-target selection on the new initializations, $m^\ast$ rises from 161 to 271 evaluations, between $3.8p$ and $8.9p$, while $p(p+1)/2$ rises from 171 to 2628. The assumed count exceeds every bootstrap replicate of $m^\ast$ at $n = 4$, $5$ and $6$, and under it QNG's budget-capped median at the strict target equals the budget in those configurations (1728, 2700 and 3888 evaluations, against Adam's 1440, 2600 and 3744). At $n = 3$ the assumed count lies inside the 95\% interval of $m^\ast$ (91--197), so the bootstrap does not decide the comparison there, although QNG's median under that count, 828, is above Adam's 792. The original initializations, on which the settings were selected, and the break-even price for final loss show the same slow growth of $m^\ast$ against the assumed count. At the best original setting, on the original initializations, the break-even prices are lower, from 64 to 123 evaluations ($1.4p$ to $3.8p$).

On the local-cost family the break-even price and the paired counts lead to different conclusions under the assumed count. With the strict-target selection on the new initializations, $m^\ast$ exceeds $p(p+1)/2$ at $n = 4$ and $5$, so QNG's budget-capped median at the strict target stays below Adam's there under the assumed count (504 against 608 and 750 against 780 evaluations). The paired comparison at the strict target under the same count, given in the table note, does not favor QNG (sign test $p = 0.25$). On final loss $m^\ast$ is at or above the assumed count in at least 99\% of the replicates at $n = 3$, $4$ and $5$, and QNG ends at the lower final loss from 57 of the 80 local-cost initializations under that count (sign test $p = 1.8\times10^{-4}$). At the best original setting the local-cost family has no break-even price, because QNG's median is not below Adam's even at the simulator price.

\section{Budgets and per-step charges at fixed settings}
\label{app:budget}

Table~\ref{tab:app-budget} gives the runs at the longer budgets of Section~\ref{sec:setup} and the quantities that change with the budget, and the last paragraph charges each step for the loss recorded after it within the same budget.

\begin{table}[tbp]
\centering
\caption{Longer and shorter budgets, QNG's metric at its simulator price (definitions in Section~\ref{sec:setup}). (a)~Settings selected on the extended grid for $\tau = 0.01$ at $B$ on the original initializations, held fixed at every budget. Counts give the initializations from which QNG or SPSA is first, Adam is first and, at a target, the two tie. At a target, first means fewer circuit evaluations, a miss counting as the budget, and on final loss the lower final loss. Each first-to-target count is printed once because it does not change with the budget. $\rho$ is QNG's or SPSA's median over Adam's (final loss, or budget-capped at a target), pooled over configurations with different budgets (972--3888 at $B$ global-cost, 648--1296 local-cost), so these medians mix budgets and need not equal Kaplan--Meier medians. SPSA entries in (a) and (b) use the shared sequence except the parenthesized 39--41, which uses the per-initialization median of five sequences. The global-cost count at $B$ switches from SPSA to Adam between the two, and from $2B$ SPSA's pooled global-cost final loss is above Adam's. (b)~Runs reaching the target on the original grid and initializations at the best original setting of Section~\ref{sec:setup}. Reach within $B/2$ is reach at a budget of $27p$.}
\label{tab:app-budget}
\small\setlength{\tabcolsep}{5pt}
\begin{tabular}{@{}lcccccc@{}}
\toprule
\textit{(a)} Strict-target settings & \multicolumn{3}{c}{Count} & \multicolumn{3}{c}{Pooled ratio $\rho$} \\
\cmidrule(lr){2-4}\cmidrule(l){5-7}
Comparison & $B$ & $2B$ & $4B$ & $B$ & $2B$ & $4B$ \\
\midrule
\multicolumn{7}{@{}l}{\textit{New initializations (20 per configuration), global-cost family}} \\
QNG--Adam, first to $\tau = 0.20$ & \multicolumn{3}{c}{12--68--0} & $1.541$ & $1.541$ & $1.541$ \\
QNG--Adam, first to $\tau = 0.01$ & \multicolumn{3}{c}{77--3--0} & $0.504$ & $0.481$ & $0.481$ \\
QNG--Adam, lower final loss & 79--1 & 79--1 & 62--18 & $8.0\times10^{-3}$ & $5.3\times10^{-3}$ & $1.2\times10^{-2}$ \\
SPSA--Adam, lower final loss & \begin{tabular}[t]{@{}c@{}}46--34\\(39--41)\end{tabular} & 1--79 & 0--80 & $0.925$ & $7.96$ & $631$ \\
\addlinespace
\multicolumn{7}{@{}l}{\textit{New initializations (20 per configuration), local-cost family}} \\
QNG--Adam, first to $\tau = 0.20$ & \multicolumn{3}{c}{36--43--1} & $1.021$ & $1.021$ & $1.021$ \\
QNG--Adam, first to $\tau = 0.01$ & \multicolumn{3}{c}{79--1--0} & $0.274$ & $0.274$ & $0.274$ \\
QNG--Adam, lower final loss & 80--0 & 79--1 & 80--0 & --- & --- & --- \\
SPSA--Adam, lower final loss & 80--0 & 75--5 & 0--80 & $3.1\times10^{-2}$ & $0.228$ & $23.1$ \\
\addlinespace
\multicolumn{7}{@{}l}{\textit{Original initializations (10 per configuration), global-cost family}} \\
QNG--Adam, first to $\tau = 0.20$ & \multicolumn{3}{c}{1--39--0} & $1.466$ & $1.466$ & $1.466$ \\
QNG--Adam, first to $\tau = 0.01$ & \multicolumn{3}{c}{39--1--0} & $0.391$ & $0.391$ & $0.391$ \\
QNG--Adam, lower final loss & 40--0 & 40--0 & 33--7 & $8.1\times10^{-3}$ & $1.5\times10^{-3}$ & $4.5\times10^{-4}$ \\
SPSA--Adam, lower final loss & 23--17 & 3--37 & 0--40 & $1.004$ & $5.86$ & $766$ \\
\addlinespace
\multicolumn{7}{@{}l}{\textit{Original initializations (10 per configuration), local-cost family}} \\
QNG--Adam, first to $\tau = 0.20$ & \multicolumn{3}{c}{17--21--2} & $1.021$ & $1.021$ & $1.021$ \\
QNG--Adam, first to $\tau = 0.01$ & \multicolumn{3}{c}{38--2--0} & $0.305$ & $0.305$ & $0.305$ \\
QNG--Adam, lower final loss & 40--0 & 40--0 & 39--1 & --- & --- & --- \\
SPSA--Adam, lower final loss & 40--0 & 37--3 & 0--40 & $3.1\times10^{-2}$ & $0.266$ & $35.5$ \\
\bottomrule
\end{tabular}
\par\vspace{10pt}
\begin{tabular}{@{}llcccccc@{}}
\toprule
\multicolumn{2}{@{}l}{\textit{(b)} Original grid, best original setting} & \multicolumn{2}{c}{Adam} & \multicolumn{2}{c}{QNG} & \multicolumn{2}{c}{SPSA} \\
\cmidrule(lr){3-4}\cmidrule(lr){5-6}\cmidrule(l){7-8}
Target & Runs per method & $B/2$ & $B$ & $B/2$ & $B$ & $B/2$ & $B$ \\
\midrule
$\tau = 0.01$ & All eight configurations (80) & 12 & 73 & 19 & 49 & 20 & 44 \\
 & Global-cost family (40) & 0 & 35 & 16 & 38 & 11 & 22 \\
\addlinespace
$\tau = 0.02$ & All eight configurations (80) & 22 & 80 & 22 & 58 & 32 & 53 \\
 & Global-cost family (40) & 4 & 40 & 17 & 38 & 16 & 27 \\
\addlinespace
$\tau = 0.20$ & All eight configurations (80) & 80 & 80 & 45 & 74 & 62 & 73 \\
\bottomrule
\end{tabular}
\par\vspace{6pt}
\begin{minipage}{\textwidth}\small A dash marks a local-cost QNG final-loss ratio, which is not given because QNG's pooled median final loss there is at floating-point zero ($1.0\times10^{-15}$, $5.6\times10^{-17}$ and $0$ at $B$, $2B$ and $4B$ against Adam's $4.9\times10^{-3}$, $2.6\times10^{-4}$ and $1.1\times10^{-6}$ on the new initializations; $6.1\times10^{-16}$, $2.8\times10^{-16}$ and $0$ against $5.1\times10^{-3}$, $2.6\times10^{-4}$ and $9.0\times10^{-7}$ on the original ones).\end{minipage}
\end{table}%

A longer budget cannot change a paired first-to-target count of block~(a). A run at $2B$ or $4B$ continues the run at $B$ with the same settings and initialization, so every crossing within $B$ is unchanged and any new crossing falls beyond $54p$ evaluations; QNG's 18th step ends at $54p + 18$ and Adam's 28th at $56p$. No QNG--Adam pair has both runs missing the target at $B$, on either family, initialization set or target, so in every pair at least one run reaches the target within $B$, before any new crossing of the other run. A tie at $B$ is the one case that a longer budget could resolve, when one run reaches the target with its last step and the other misses. No global-cost pair is tied at either target, and the only tied pairs are local-cost pairs at the loose target that start at or below the target, so both runs cross at zero evaluations under every budget.

What changes with the budget is final loss, the values charged to runs that miss the target within $B$, and SPSA's standing against Adam. At $B$, under the strict-target selection, QNG ends the budget with the lower final loss from 79 of the 80 new global-cost initializations, with a median paired difference of $-2.04$ in $\log_{10}$ final loss (95\% interval $-2.38$ to $-1.76$), and at the best original setting it ends lower from 38 of the 40 original global-cost initializations. On the new global-cost initializations QNG's final-loss lead over Adam narrows at $4B$, from 79 to 62 of the 80 initializations, and QNG still ends lower in most pairs of every configuration (19, 13, 16 and 14 of 20 at $n = 3$ to $6$). The pooled ratios at a target move only through the values charged to runs that miss the target within $B$. SPSA's final-loss comparison with Adam reverses with the budget. On the local-cost family SPSA ends lower from all 80 new initializations at $B$ and from none at $4B$. On the global-cost family, where the count at $B$ depends on SPSA's perturbation sequence and no ordering is stated, Adam ends lower from 79 of 80 at $2B$ and from all 80 at $4B$.

Reach rates, unlike the paired counts of block~(a), depend on the budget at which they are evaluated, because a run that misses the target within one budget can reach it within a longer one. In block~(b), at $\tau = 0.01$, the order of the three methods' reach reverses completely, from Adam, QNG, SPSA within $B$ to SPSA, QNG, Adam within $B/2$. On the global-cost family no Adam run reaches $\tau = 0.01$ within $B/2$, while QNG reaches it in 16 of 40. In the rows at $\tau = 0.02$, the examined target closest to the strict one (Section~\ref{sec:setup}), the reach order changes in the same direction, from Adam, QNG, SPSA within $B$, with 80, 58 and 53 runs, to SPSA first within $B/2$, with 32 runs against 22 for each of the other two methods. At the loose target Adam reaches the target in every run within either budget.

Charging the loss recorded after each step changes neither the gap between the two readings of Section~\ref{sec:audit} nor the main-comparison counts of Section~\ref{sec:target}. A step then costs one evaluation more, $2p + 1$ for Adam, $3p + 2$ for QNG and $3$ for SPSA, within the same budget $B$, so Adam keeps 25 or 26 of its 27 steps, QNG all 17 and SPSA $18p$ of its $27p$. On the original grid at the loose target, SPSA's pooled ratio to Adam under its shared sequence is 1.35 (0.89--1.86) over reached runs only and 2.82 (2.21--3.34) with misses charged the budget on the original initializations, and 1.12 (0.96--1.47) and 2.92 (2.20--3.32) on the new ones, with intervals adjusted as in Appendix~\ref{app:conv}. The conditional interval still covers 1 and the budget-capped interval lies above it on both sets, and QNG's intervals lie above 1 under both readings. Within configurations the conditional reading leaves SPSA slower than Adam in 7 of the 8 configurations on the original initializations and in all 8 on the new ones. In the main comparison QNG is first from 12 of the 80 new global-cost initializations at the loose target and Adam from 68, and at the strict target QNG is first from 77 and Adam from 3, the same counts as without the charge. Adam's reach of the strict target falls from 65 to 59 of 80, because the six of its runs that reached it at exactly $B$ now miss it, and QNG is first in each of these six pairs with or without the charge.

\section{A second protocol}
\label{app:prot2}

The second protocol of Section~\ref{sec:target} runs Adam, QNG and SPSA with one fixed setting each and a budget of 2000 circuit evaluations per run, a miss being charged 2000, from 48 initializations per configuration that the three methods share. The fixed settings are $\eta = 0.1$ for Adam, $\eta = 0.2$ with $\lambda = 10^{-3}$ for QNG, and a gain of 0.1 with a perturbation size of 0.1 for SPSA. The protocol evaluates time to target at its own targets, 0.55 on the global-cost family and 0.20 on the local-cost family. Its reference arm uses the circuits and cost families of Section~\ref{sec:setup}, and its parameter-tied arm uses a different circuit in which every layer repeats one layer's $2n$ parameters.

At its own targets QNG has the lowest budget-capped median in no configuration of either arm. In the reference arm Adam has the lowest in the four global-cost configurations and SPSA in the four local-cost configurations, and Adam reaches the target before QNG from 189 of the 192 global-cost and 186 of the 192 local-cost initializations. In the parameter-tied arm SPSA has the lowest budget-capped median in seven of the eight configurations and Adam in the remaining one.

Evaluated at the strict target $\tau = 0.01$ from the same trajectories, the reference arm reverses the verdict on the global-cost family. QNG reaches the target first from 119 of the 192 global-cost initializations and Adam from none, and neither method reaches it from the other 73 (sign test $p = 3.0\times10^{-36}$). QNG's budget-capped median is below Adam's at $n = 3$ and $4$ (495 against 1044 and 1067 against 1824 evaluations), and both medians equal the budget at $n = 5$ and $6$. On the local-cost family Adam remains first from 178 of 192, the opposite of the local-cost result of Appendix~\ref{app:select}, in which QNG is first from 79 of 80. One possible reason, which we have not tested, is that this protocol fixes each method's setting while Section~\ref{sec:target} selects it for the strict target. In the parameter-tied arm the global-cost count at the strict target is 52 for QNG against 67 for Adam (sign test $p = 0.199$), so the reversal does not appear there. The second protocol therefore corroborates the reversal of Section~\ref{sec:target} on the global-cost family of this paper's circuit and does not extend it to a different circuit.

\section{Pooling across configurations}
\label{app:pool}

Pooling across configurations, which makes every pooled median mix budgets (Section~\ref{sec:setup}), also changes verdicts. Section~\ref{sec:audit} gives one instance, and this appendix gives two more. A median pooled across configurations can hide a lead present in every configuration, and it can reverse while every per-configuration result stays fixed.

At the best original setting (Adam $\eta = 0.3$; QNG $\eta = 0.3$, $\lambda = 10^{-3}$) and the strict target, pooling the two cost families hides a QNG lead present in every global-cost configuration. On both initialization sets QNG's budget-capped median at $\tau = 0.01$ is below Adam's in each global-cost configuration. On the original initializations these medians are the global-cost QNG and Adam entries of the strict-target block of Table~\ref{tab:app-cells}, where both methods' best setting is the best original setting. On the new initializations they are 385 against 792, 824.5 against 1440, 1661 against 2600 and 2712.5 against 3744 evaluations at $n = 3$ to $6$. Pooled over the global-cost family, QNG's median is 970 against Adam's 1832 on the original initializations and 1208 against 1764 on the new ones. Pooled over the local-cost family it is 920 against 604 and 920 against 672, and pooled over all eight configurations it is 970 against 908 and 1080 against 920, so the pooled reading puts Adam ahead on both sets. QNG's local-cost medians at $n = 4$, $5$ and $6$ equal their configurations' budgets on both sets and carry no time estimate.

With the global-cost settings selected for the loose target (Table~\ref{tab:app-grid}) held fixed on the original initializations, the ratio of QNG's to Adam's pooled median final loss moves from 0.56 at $B$ to 2.2 at $2B$ and 43 at $4B$, while the method with the lower median final loss is the same in every configuration at every budget. QNG's 20 lowest final losses are its runs at $n = 3$ and $4$ at every budget, so its 40 final losses form two clusters and the pooled median falls between them. In the paired comparison QNG ends lower from 22, 20 and 22 of the 40 initializations. The instance depends on QNG's damping of $10^{-2}$ at $n = 5$ and $6$ under this selection, and it disappears under the strict-target selection, where the pooled final-loss ratio stays below 1 at every budget (Table~\ref{tab:app-budget}).

\section{Further related work}
\label{app:related}

The benchmarking practices of Section~\ref{sec:related} charge a miss in different ways. Performance profiles give a solver that fails on a problem a penalty value instead of discarding the problem \citep{dolan2002}, and \citet{more2009} set the time of a run that never satisfies the convergence test to infinity. The expected run time of \citet{hoos1998} charges every unsuccessful run its cutoff time, the average runtime of the COCO platform counts the evaluations of unsuccessful trials \citep{hansen2016}, and a benchmark of neural-network training assigns a run that misses its target an infinite training time \citep{dahl2023}. An analysis of the competition run on that benchmark notes that charging the full time budget for a workload that a submission fails to train is the most favorable assumption for that submission \citep[Sec.~3.1]{kasimbeg2025}, and the charge $B$ of Appendix~\ref{app:conv} is a lower bound in the same sense. The same literature compares solvers at several accuracy levels, because the target sets the difficulty and the character of the problem being solved \citep{hansen2016} and because users need both low-accuracy and high-accuracy solutions \citep{more2009}. A survey of benchmarking practice in optimization notes that the better algorithm can depend on the chosen time budget or target value, and that a comparison at a fixed target has to state how it treats runs that never reach the target \citep[Sec.~5]{bartzbeielstein2020}.

Comparisons of variational quantum optimizers also condition summaries on success without timing the runs. One comparison keeps only the best half of its runs \citep{pellowjarman2021}, another counts a starting point as converged when any of ten repeated runs from it converges \citep{gacon2021}, and a benchmark of natural-gradient and second-order preconditioning estimates Lipschitz constants only from the runs that reach a 1\% tolerance \citep[Sec.~3.2]{lisartliebermann2025}.

\citet[Sec.~4.4]{beiranvand2017} state that tuning some methods while leaving the others at their default settings is not an appropriate comparison, and \citet[Sec.~2]{birattari2006} require that the problems used to assess an algorithm not be used to develop it. \citet[Sec.~4.2 and App.~E]{choi2019} report that their conclusions about inclusion relationships between optimizers do not depend on the step budgets and error targets they chose, and they note that large changes to these values would require new experiments.

\citet[Secs.~III and VI]{bonetmonroig2023} compare four gradient-free optimizers by their median final relative energy error at fixed total budgets of measurement shots, and they find that the preferred optimizer depends on the problem and on the budget, with CMA-ES preferable in a high-cost, high-accuracy setting. \citet[Sec.~6]{lavrijsen2020} give each optimizer a budget of objective-function calls and report the average and the minimum of the final values. These studies evaluate the loss at the end of the budget, and the comparisons cited in Section~\ref{sec:intro} time only the runs that reached a target. We compute time to target both ways on one set of runs of Adam, exact-metric QNG and SPSA in exact, noiseless simulation. \citet[Sec.~3.1]{gacon2021} approximate the metric because on hardware the full object generally requires a number of function evaluations quadratic in the number of parameters. Keeping the metric exact removes estimation error as an explanation of QNG's verdicts.

\citet[Sec.~3]{stokes2020} find QNG ahead of Adam per iteration from a single initialization, with block-diagonal and diagonal approximations of the metric. \citet[Sec.~3 and Table~1]{fitzek2024} give all methods one shared step size and find block-diagonal QNG ahead of Adam within the first 5000 evaluations, while Adam needs fewer evaluations to reach the tight target of a 0.99 approximation ratio at every depth they report. \citet[Fig.~5(a)]{haug2021b} find QNG behind Adam in early iterations and ahead of it after more iterations. The two changes of order run in opposite directions. A benchmark of a quantum annealer also reports an order in samples to target that changes with the target, with one solver worse than two others at hitting the hardest targets and better at hitting the easiest \citep[Sec.~5]{king2015}. Since optimization benchmarking already evaluates comparisons at several targets (Section~\ref{sec:related}), the open question is a controlled test of such a change.

\citet{wierichs2020} find that natural gradient with the full metric fails less often than BFGS and Adam in exact simulation, and they also price that metric. Their comparison covers minimal and overparameterized variational circuits for spin models, with learning rates chosen by hand and without a matched evaluation budget (their Secs.~II.E and III). They estimate the quantum cost of natural gradient to be considerably higher than that of BFGS and Adam where those methods converge (their Sec.~III.A), while their published abstract presents the total run times of natural gradient as competitive. They stopped the Adam runs that failed on the minimal circuit at $5 \cdot 10^4$ iterations and do not exclude convergence with more iterations. Their question is how often each optimizer reaches the minimum. We ask how a comparison at a fixed evaluation budget weighs the same trade-off between reliability and price under each reporting convention.

\Citet{vanstraaten2021} analyze the measurement cost of metric-aware variational algorithms and give the conditions under which paying for the metric is worthwhile. In their accounting, the share of the measurement cost spent on the metric becomes asymptotically negligible as the numbers of iterations and qubits grow (their Secs.~III--IV). Their currency is the number of repeated measurements (shots) on hardware, while this paper counts exact circuit evaluations in simulation and charges the metric its simulator price. Because this paper does not measure the hardware price of the metric, Section~\ref{sec:target} reports the largest charge per metric at which QNG keeps its strict-target advantage over Adam.

Benchmarking practice is itself under scrutiny in quantum machine learning. \citet{bowles2024} argue that experimental-design choices strongly shape the results of simulated benchmarks of quantum models, and they find that across the classification tasks of their study standard off-the-shelf classical models do better overall than the quantum classifiers. \citet{schuld2022} ask whether outperforming classical machine learning should remain the dominant goal of research in quantum machine learning. We place our study in that line of work and apply it to optimizers rather than to models. Its primary objects are two reporting conventions; the verdict between optimizers makes their effect visible. The barren-plateau background it relies on is standard \citep{mcclean2018, larocca2025}. In a barren plateau the Hessian entries are also exponentially suppressed \citep{cerezo2021b}, and so are the loss differences on which gradient-free optimizers base their steps \citep{arrasmith2021}.

\section{The exact metric along the depth ladder}
\label{app:ladder}

The ladder of Section~\ref{sec:ladder} measures the metric separately from the optimizer runs. It takes the global-cost circuit at $n = 6$ qubits with $L = 1, \dots, 6$ layers, so the parameter count grows from $p = 12$ to $p = 72$. Its deepest rung is the global-cost configuration with $n = 6$ of Sections~\ref{sec:audit} and~\ref{sec:target}; the shallower rungs are not optimizer configurations. Three summaries describe the metric $g$ at each depth. The condition number $\kappa(g)$ measures how unevenly $g$ scales different directions of parameter space. The effective rank compares the trace of $g$ with its largest eigenvalue $\lambda_{\max}$, which is a different quantity from the damping $\lambda$; the effective rank equals $p$ when all eigenvalues are equal and approaches $1$ when one eigenvalue dominates. It is a participation-style summary and differs from the participation ratio, which divides the squared sum of the eigenvalues by the sum of their squares. It also differs from the effective dimension of \citet{abbas2021}, which integrates a determinant built from the normalized Fisher information over the parameter space at a resolution set by the number of data samples. The alignment is the cosine between the gradient and the damped natural-gradient direction $(g + \lambda I)^{-1}\nabla C$, and it equals $1$ when the preconditioner leaves the direction of the step unchanged. Table~\ref{tab:spectrum} gives the three summaries and the true rank at each depth. Because the measurement keeps only means over random initial points, it describes the depth trend and attaches no uncertainty to it.

As depth grows, the condition number of the metric rises steeply and not monotonically, while its effective rank stays between $8.0$ and $10.5$ and its true rank grows with the parameter count. From $L = 1$ to $L = 6$ the mean $\log_{10}\kappa(g)$ rises from $1.87$ to $5.22$ and peaks at $5.44$ at $L = 5$. Over the same depths the true rank grows from $12$ to $72$, so the effective rank does not keep pace with the size $p$ of the matrix being inverted. \Citet[Sec.~IV]{haug2021} count the independent directions in which a circuit can move its state by the rank of this metric, so that a redundant parameter adds a zero eigenvalue. The metric here has full rank at every depth, and its low effective rank comes from eigenvalues of very unequal size, not from a missing direction. For layered circuits of random single-qubit rotations and CNOT gates on 10 qubits, \citet[Sec.~V, Fig.~3]{haug2021} find that this rank grows linearly with the number of parameters until it saturates at $2(2^n - 1)$, and that the variance of the logarithms of the nonzero eigenvalues peaks at that saturation point; \citet{larocca2023} bound the maximal rank of the quantum Fisher information by the dimension of the dynamical Lie algebra of the circuit's generators. With $p \le 72$ against a ceiling of 126 at $n = 6$, the ladder stays below that saturation. Relative to the rank results of \citet{haug2021}, we add $\kappa(g)$ and the effective rank of a full-rank metric along a ladder whose deepest rung is an optimizer configuration.

The alignment falls from $1.000$ at $L = 1$ to $0.294$ at $L = 6$, and it is not monotone either, since it is $0.434$ at $L = 2$ and $0.436$ at $L = 3$. At $L = 1$ the alignment is $1.000$ although $\log_{10}\kappa(g)$ is already $1.87$. In each layer every qubit receives an $R_Y$ rotation followed by an $R_Z$ rotation (Section~\ref{sec:setup}). With one layer the gradient of the global cost has no component along the $R_Z$ angles, and the metric acts on the $R_Y$ angles as a multiple of the identity without coupling them to the $R_Z$ angles. The preconditioner therefore rescales the gradient without turning it, while each QNG step at this depth is still charged $1.54$ times the circuit evaluations of an Adam step. The ladder fixes the damping at $\lambda = 10^{-3}$ and evaluates the metric at random initial points. The alignment averaged over each QNG run on the original grid is a different quantity, which depends on the damping and on the trajectory; its median over the runs at each damping is $0.65$ at $\lambda = 10^{-3}$ and $0.94$ at $\lambda = 10^{-1}$.

We offer no mechanism for the verdicts of Sections~\ref{sec:audit} and~\ref{sec:target} and no per-run predictor. At the deepest rung of the ladder, QNG at its best original setting on the original initializations has a budget-capped median above Adam's at the loose target and below Adam's at the strict target (Table~\ref{tab:app-cells}), so one spectrum accompanies a deficit at one target and a lead at the other. Whether a feature of the metric at initialization predicts which method reaches the target first from a given initialization cannot be tested on these runs. The second protocol of Appendix~\ref{app:prot2}, in which each initialization gives one run of each method, would supply such a test, but QNG reaches that protocol's own target first from only 1 of its 192 global-cost and 4 of its 192 local-cost initializations, so the question remains open in both directions.

\end{document}